\documentclass{article} %
\usepackage{iclr2027_conference,times}

\usepackage{amsmath,amsfonts,bm}

\def\eqref#1{equation~\ref{#1}}

\def\1{\bm{1}}

\DeclareMathAlphabet{\mathsfit}{\encodingdefault}{\sfdefault}{m}{sl}
\SetMathAlphabet{\mathsfit}{bold}{\encodingdefault}{\sfdefault}{bx}{n}

\usepackage{hyperref}
\usepackage{url}
\usepackage{amsmath,amssymb,mathtools}
\usepackage{booktabs,multirow,colortbl,array}
\usepackage{graphicx,subcaption}
\usepackage{wrapfig}
\usepackage{caption}
\usepackage{afterpage}
\usepackage{placeins} %
\usepackage{needspace}
\usepackage{xcolor}
\colorlet{blue}{black}
\hypersetup{hidelinks}
\usepackage{tikz} %
\usepackage{xspace}
\usepackage{enumitem}
\usepackage{algorithm}
\usepackage{algpseudocode}

\newcommand{\method}{\textsc{FlipDir}\xspace}
\newcommand{\bench}{\textsc{VisFlip}\xspace}
\newcommand{\ours}{\textbf{\textsc{FlipDir}}\xspace}

\newcommand{\hrec}{H(\mathrm{C},\mathrm{N},\mathrm{F})}
\newcommand{\dH}[1]{\,{\scriptsize\textcolor{green!45!black}{(+#1)}}}   %

\title{Looks the Same, Answers Differently: \\ Flip-Direction Steering for Robust Vision-Language Reasoning}

\author{\normalfont
\begin{minipage}{\dimexpr\textwidth-2\tabcolsep\relax}
\centering
Yeonsung Jung$^{1}$, Joonhyun Jeong$^{1,2}$, Hoang Pham$^{3,4}$,\\
Joowon Kim$^{1}$, Yoonsik Park$^{1}$, Viet Dac Lai$^{5,\dagger}$, Eunho Yang$^{1,6,\dagger}$\\[0.6em]
{\small
$^{1}$Graduate School of AI, KAIST \quad $^{2}$NAVER Cloud\\
$^{3}$Department of Computer Science and Engineering, The Ohio State University\\
$^{4}$Department of Biomedical Informatics, The Ohio State University\\
$^{5}$Adobe Research \quad $^{6}$AITRICS}
\end{minipage}}

\iclrfinalcopy %
\begin{document}
\raggedbottom

\maketitle
\lhead{Preprint}
\begingroup
\renewcommand{\thefootnote}{\fnsymbol{footnote}}
\footnotetext[2]{Co-corresponding authors.}
\endgroup

\begin{abstract}
{Vision-language models (VLMs) achieve strong visual reasoning performance, yet subtle changes from routine image capture and processing can alter their reasoning trajectories even when images appear nearly identical. In long-horizon generation, the resulting activation shifts may accumulate across decoding steps, progressively altering reasoning tokens and ultimately changing the final answer, a phenomenon referred to as \emph{answer flips}. To address this instability, we propose \method (Flip-Direction Steering), a training-free inference-time method that estimates a low-rank flip-inducing activation subspace from contrastive pairs of original and answer-flipping inputs and selectively steers hidden states during decoding. A margin-based gate limits subspace attenuation to uncertain decoding steps, recovering original predictions while preserving stable ones. To evaluate robustness beyond accuracy or consistency on fixed test sets, we introduce \bench, a benchmark framework that constructs evaluation groups for a  target model and visual variation setting to separately assess recovery of original predictions and preservation of stable ones. \bench spans nine dataset--variation combinations across scientific reasoning, robot-scene understanding, and medical VQA, covering subtle visual variations common in each domain. Experiments across 18 settings demonstrate that \method consistently outperforms existing methods on the combined recovery and preservation metric.} We will make our code publicly available.
\end{abstract}

\setlength{\parskip}{2pt}
\makeatletter
\renewcommand{\paragraph}{\@startsection{paragraph}{4}{\z@}{3pt}{-1em}{\normalfont\normalsize\bfseries}}
\makeatother

\afterpage{\afterpage{
\begin{figure}[!t]
\centering
\includegraphics[width=\linewidth]{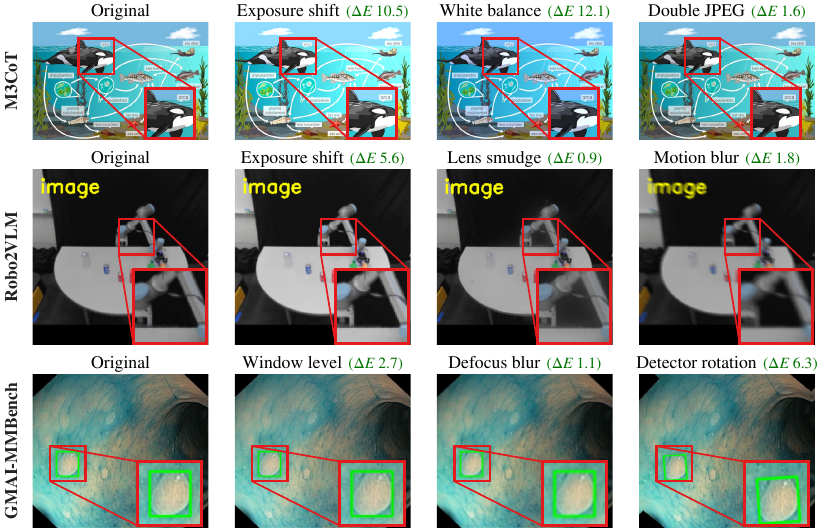}
\caption{\textbf{Visual variations in \bench.} Rows show subtle variations common to each domain. $\Delta E$ denotes the mean per-pixel color difference from the original image (parameter ranges in Appendix~\ref{app:variation-details}).}
\label{fig:variations}
\end{figure}
\afterpage{
\begin{figure}[!t]
\centering
\includegraphics[width=\linewidth]{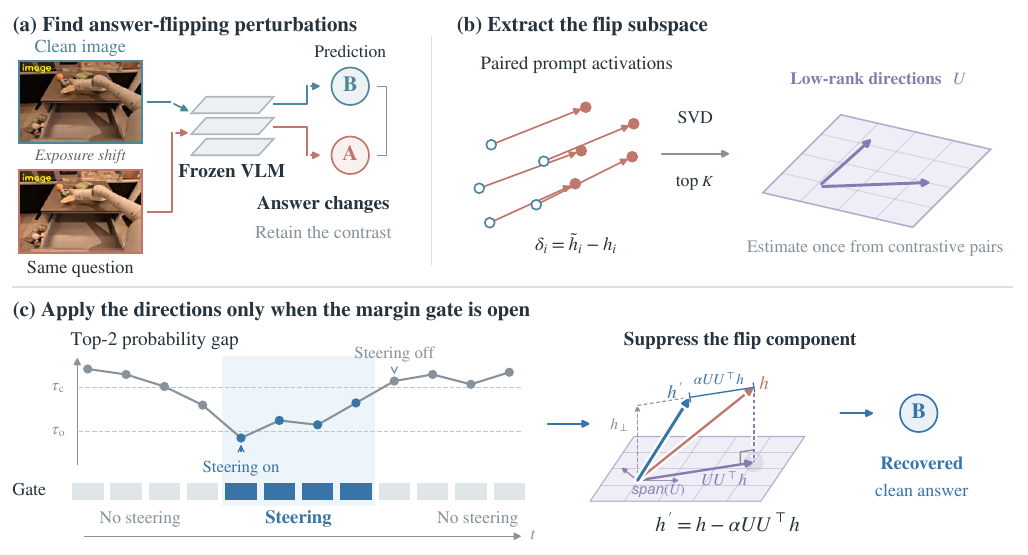}
\caption{\textbf{\method overview.} (a) Identify \emph{answer flips}. (b) Estimate $U$ via SVD of paired activation differences. (c) Attenuate the subspace projection at margin-gated decoding steps. The plane denotes $\mathrm{span}(U)$, and increasing $\alpha$ moves $h'$ toward $h_{\perp}=(I-UU^{\top})h$.}
\label{fig:method-overview}
\end{figure}
\afterpage{\afterpage{\afterpage{
\begin{table}[!t]
\centering
\caption{Main results on \textbf{Qwen3-VL-8B} (\%): clean preservation (C), non-flip preservation (N), flip recovery (F), and their harmonic mean H. Best H is bold; green differences indicate \ours{}'s gain in H over the strongest baseline.}
\label{tab:main-qwen}
\footnotesize
\renewcommand{\arraystretch}{0.92}
\setlength{\tabcolsep}{2pt}
\begin{tabular*}{\linewidth}{@{\extracolsep{\fill}}l ccc>{\columncolor{gray!10}}c ccc>{\columncolor{gray!10}}c ccc>{\columncolor{gray!10}}c@{}}
\toprule
\emph{M3CoT} & \multicolumn{4}{c}{Exposure} & \multicolumn{4}{c}{White balance} & \multicolumn{4}{c}{JPEG} \\
\cmidrule(lr){2-5}\cmidrule(lr){6-9}\cmidrule(lr){10-13}
Method & C & N & F & H & C & N & F & H & C & N & F & H \\ 
\midrule
Base   & 100 & 100 & 0.0 & 0.0 & 100 & 100 & 0.0 & 0.0 & 100 & 100 & 0.0 & 0.0 \\ 
VTI    & 54.1 & 50.5 & 33.9 & 44.2 & 54.1 & 56.3 & 37.9 & 47.9 & 54.1 & 57.7 & 38.5 & 48.6 \\ 
VCD    & 81.6 & 84.6 & 60.7 & 74.0 & 83.0 & 82.7 & 62.6 & 74.8 & 82.3 & 82.5 & 55.6 & 71.0 \\ 
LEAD   & 87.6 & 89.0 & 62.8 & 77.8 & 87.6 & 87.6 & 56.3 & 73.9 & 87.6 & 87.4 & 56.1 & 73.8 \\ 
\ours  & 87.4 & 87.9 & 64.5 & \textbf{78.3}\dH{0.5} & 84.6 & 84.1 & 64.2 & \textbf{76.4}\dH{1.6} & 87.2 & 87.1 & 66.3 & \textbf{78.9}\dH{5.1} \\ 
\midrule
\emph{Robo2VLM} & \multicolumn{4}{c}{Exposure} & \multicolumn{4}{c}{Lens smudge} & \multicolumn{4}{c}{Motion blur} \\
\cmidrule(lr){2-5}\cmidrule(lr){6-9}\cmidrule(lr){10-13}
Base   & 100 & 100 & 0.0 & 0.0 & 100 & 100 & 0.0 & 0.0 & 100 & 100 & 0.0 & 0.0 \\ 
VTI    & 90.2 & 90.0 & 61.7 & 78.1 & 90.3 & 89.9 & 54.7 & 74.1 & 90.2 & 90.7 & 55.3 & 74.6 \\ 
VCD    & 91.1 & 90.5 & 62.6 & 79.0 & 91.9 & 93.0 & 66.3 & 81.7 & 91.1 & 89.4 & 63.8 & 79.3 \\ 
LEAD   & 96.1 & 94.3 & 60.7 & 80.1 & 96.4 & 97.1 & 58.1 & 79.2 & 96.1 & 94.1 & 47.9 & 71.6 \\ 
\ours  & 92.5 & 93.2 & 67.3 & \textbf{82.4}\dH{2.3} & 94.1 & 94.0 & 66.3 & \textbf{82.5}\dH{0.8} & 93.2 & 93.0 & 67.0 & \textbf{82.4}\dH{3.1} \\ 
\midrule
\emph{GMAI-MMBench} & \multicolumn{4}{c}{Window level} & \multicolumn{4}{c}{Defocus blur} & \multicolumn{4}{c}{Rotation} \\
\cmidrule(lr){2-5}\cmidrule(lr){6-9}\cmidrule(lr){10-13}
Base   & 100 & 100 & 0.0 & 0.0 & 100 & 100 & 0.0 & 0.0 & 100 & 100 & 0.0 & 0.0 \\ 
VTI    & 45.8 & 47.4 & 39.3 & 43.9 & 45.8 & 43.4 & 32.9 & 39.9 & 45.8 & 42.7 & 37.3 & 41.6 \\ 
VCD    & 71.4 & 78.9 & 59.8 & 69.1 & 76.6 & 72.5 & 45.9 & 61.7 & 72.4 & 76.0 & 50.0 & 63.9 \\ 
LEAD   & 85.4 & 81.1 & 45.5 & 65.2 & 85.4 & 78.8 & 40.0 & 60.7 & 85.4 & 84.4 & 44.5 & 65.2 \\ 
\ours  & 79.7 & 84.2 & 54.5 & \textbf{70.1}\dH{1.0} & 80.2 & 81.5 & 44.7 & \textbf{63.7}\dH{2.0} & 80.2 & 81.8 & 56.4 & \textbf{70.7}\dH{5.5} \\ 
\bottomrule
\end{tabular*}
\vspace{0.4em}
\centering
\caption{Main results on \textbf{Gemma-3-12B}; notation as in Table~\ref{tab:main-qwen}.}
\label{tab:main-gemma}
\footnotesize
\renewcommand{\arraystretch}{0.92}
\setlength{\tabcolsep}{2pt}
\begin{tabular*}{\linewidth}{@{\extracolsep{\fill}}l ccc>{\columncolor{gray!10}}c ccc>{\columncolor{gray!10}}c ccc>{\columncolor{gray!10}}c@{}}
\toprule
\emph{M3CoT} & \multicolumn{4}{c}{Exposure} & \multicolumn{4}{c}{White balance} & \multicolumn{4}{c}{JPEG} \\
\cmidrule(lr){2-5}\cmidrule(lr){6-9}\cmidrule(lr){10-13}
Method & C & N & F & H & C & N & F & H & C & N & F & H \\ 
\midrule
Base   & 100 & 100 & 0.0 & 0.0 & 100 & 100 & 0.0 & 0.0 & 100 & 100 & 0.0 & 0.0 \\ 
VTI    & 39.2 & 34.9 & 31.7 & 35.0 & 39.2 & 38.8 & 29.0 & 35.0 & 39.2 & 35.4 & 33.9 & 36.0 \\ 
VCD    & 66.4 & 66.2 & 42.4 & 55.8 & 66.1 & 66.3 & 48.2 & 58.9 & 66.1 & 67.3 & 44.3 & 57.1 \\ 
LEAD   & 63.0 & 66.4 & 47.8 & 57.9 & 63.0 & 64.7 & 51.8 & 59.3 & 61.0 & 64.9 & 56.0 & 60.4 \\ 
\ours  & 70.2 & 69.1 & 50.0 & \textbf{61.6}\dH{3.7} & 74.4 & 72.5 & 45.2 & \textbf{60.8}\dH{1.5} & 70.6 & 70.9 & 51.7 & \textbf{63.0}\dH{2.6} \\ 
\midrule
\emph{Robo2VLM} & \multicolumn{4}{c}{Exposure} & \multicolumn{4}{c}{Lens smudge} & \multicolumn{4}{c}{Motion blur} \\
\cmidrule(lr){2-5}\cmidrule(lr){6-9}\cmidrule(lr){10-13}
Base   & 100 & 100 & 0.0 & 0.0 & 100 & 100 & 0.0 & 0.0 & 100 & 100 & 0.0 & 0.0 \\ 
VTI    & 48.3 & 46.6 & 32.9 & 41.4 & 48.3 & 48.9 & 32.7 & 41.8 & 48.3 & 49.1 & 33.2 & 42.1 \\ 
VCD    & 80.1 & 79.5 & 41.6 & 61.1 & 80.6 & 79.7 & 47.1 & 64.9 & 78.8 & 80.7 & 44.9 & 63.4 \\ 
LEAD   & 84.5 & 85.3 & 56.4 & 72.7 & 83.1 & 83.6 & 52.9 & 69.9 & 84.5 & 83.0 & 44.4 & 64.6 \\ 
\ours  & 88.2 & 87.8 & 55.0 & \textbf{73.4}\dH{0.7} & 87.8 & 86.4 & 56.2 & \textbf{73.6}\dH{3.7} & 85.7 & 82.8 & 50.8 & \textbf{69.1}\dH{4.5} \\ 
\midrule
\emph{GMAI-MMBench} & \multicolumn{4}{c}{Window level} & \multicolumn{4}{c}{Defocus blur} & \multicolumn{4}{c}{Rotation} \\
\cmidrule(lr){2-5}\cmidrule(lr){6-9}\cmidrule(lr){10-13}
Base   & 100 & 100 & 0.0 & 0.0 & 100 & 100 & 0.0 & 0.0 & 100 & 100 & 0.0 & 0.0 \\ 
VTI    & 39.5 & 41.7 & 34.3 & 38.3 & 39.5 & 42.9 & 33.8 & 38.4 & 39.5 & 41.6 & 35.7 & 38.8 \\ 
VCD    & 82.0 & 81.9 & 45.1 & 64.4 & 78.5 & 81.2 & 42.9 & 62.0 & 79.0 & 83.2 & 42.1 & 61.9 \\ 
LEAD   & 80.5 & 76.9 & 51.0 & 66.6 & 80.5 & 75.9 & 48.1 & 64.6 & 80.5 & 75.1 & 44.4 & 62.2 \\ 
\ours  & 88.5 & 89.4 & 49.0 & \textbf{70.0}\dH{3.4} & 82.5 & 84.3 & 54.5 & \textbf{70.9}\dH{6.3} & 84.5 & 83.2 & 43.7 & \textbf{64.2}\dH{2.0} \\ 
\bottomrule
\end{tabular*}
\end{table}
}}}}}}

\section{Introduction}
\label{sec:intro}

{Recent vision-language models (VLMs) achieve strong performance across scientific reasoning, robot-scene understanding, and medical visual question answering \citep{mmcot2023,robo2vlm2025,li2023llavamed}. Despite these advances, generating reliable and consistent responses to diverse visual inputs in real-world scenarios remains challenging \citep{rosenfeld2025stability}. This requires models to preserve task-relevant visual information across image variations and integrate it consistently with the textual context throughout generation. These challenges have motivated research on several aspects of VLM reliability, including adversarial robustness \citep{xie2025chainattack}, multimodal safety \citep{wang2025mml}, and visual hallucination \citep{vti2024}. Recent studies further show that, even without deliberate attacks, benign corruptions such as additive noise can degrade VLM performance \citep{mllmic2025}.}

\begin{figure}[t]
\centering
\includegraphics[width=\linewidth]{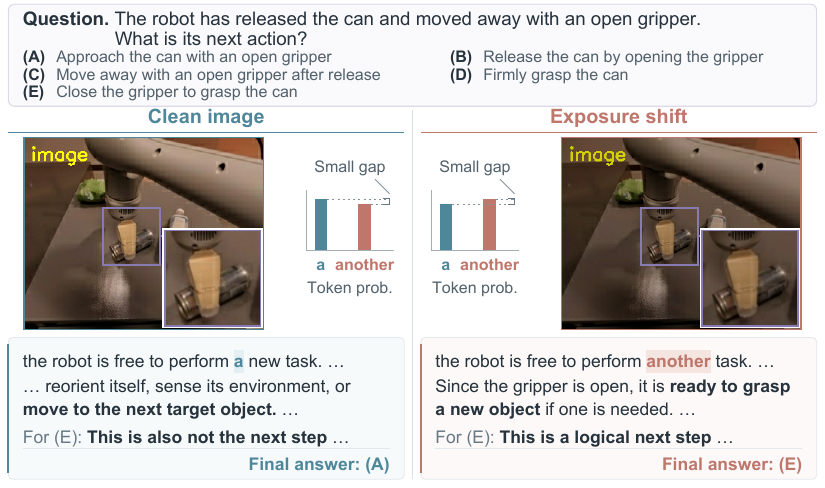}
\caption{\textbf{A subtle exposure shift can flip the final answer.} For nearly identical Robo2VLM images, Qwen3-VL-8B produces identical text before diverging at \emph{a} versus \emph{another} and reaching different action predictions. Token probabilities are schematic.}
\label{fig:qualitative}
\end{figure}

{However, common real-world scenarios involve benign inputs with subtle visual variations rather than deliberate attacks, unsafe requests, or pronounced image degradation. Although safety and resistance to adversarial prompts receive considerable attention in recent post-training \citep{gemma32025,llama42025}, the effects of subtle visual variations, such as slight exposure shifts or JPEG compression, on long-horizon visual reasoning remain underexplored. As shown in Table~\ref{tab:prevalence}, we observe that subtle exposure shifts can change final predictions even in frontier models, a phenomenon we call \emph{answer flips}. On M3CoT \citep{m3cot2024}, a benchmark for multi-step multimodal reasoning, flip rates are $9.6\%$ for GPT-5.5 and $21.7\%$ for Claude Opus 4.8 \citep{gpt552026,claude2026} (details in Sec.~\ref{sec:flipbench} and Appendix~\ref{app:kmatch}). Figure~\ref{fig:qualitative} shows an \emph{answer flip} under a subtle exposure shift, where initially identical generations first differ at \emph{a new task} versus \emph{another task} and later reach different action predictions. Moreover, these early token changes can remain locally plausible, making it difficult to identify when intervention is needed.}

\begin{table}[H]
\centering
\caption{\textbf{\emph{Answer flips} on recent models.}}
\label{tab:prevalence}
\small
\begin{tabular}{lcccc}
\toprule
 & GPT-5.5 & Claude Opus 4.8 & Qwen3-VL-8B & Gemma-3-12B \\
\midrule
Clean accuracy (\%)   & 93.2 & 91.8 & 68.3 & 47.8 \\
\emph{Answer flip} (\%) & \textbf{9.6} & \textbf{21.7} & \textbf{47.6} & \textbf{62.4} \\
\bottomrule
\end{tabular}
\end{table}

{To address this challenge, we propose \method (Flip-Direction Steering), a training-free inference-time method that estimates a low-rank flip-inducing activation subspace and uses it to selectively steer hidden states during decoding. We estimate this subspace offline by applying singular value decomposition (SVD) to activation differences from a set of contrastive pairs of original and answer-flipping inputs. During inference on new inputs, a margin-based gate uses the gap between the two largest next-token probabilities to select uncertain decoding steps for intervention (Figure~\ref{fig:margin-divergence}). When the gate opens, we attenuate the part of the current hidden state that lies in the estimated subspace; otherwise, we retain the unedited next-token distribution. Calibration requires no gradient updates or auxiliary model training, and inference uses the stored basis without a corresponding original image.}

{Our evaluation targets robustness to subtle visual variations rather than improvements in task accuracy. Existing benchmarks typically measure accuracy or consistency on fixed evaluation sets \citep{mllmic2025,mmcbench2024}. However, accuracy can increase when a visual change happens to correct an originally wrong prediction, even though the model fails to maintain its original decision. High consistency may result from perturbations that barely affect the model, while increasing their severity can make the evaluation less representative of real-world scenarios. To overcome these limitations, we introduce \bench, a benchmark framework that constructs evaluation groups for a chosen model, dataset, and visual variation setting based on whether the original predictions change. This model-specific construction enables separate evaluation of recovery and preservation in the setting of interest. We instantiate the framework across nine dataset--variation combinations spanning scientific reasoning, robot-scene understanding, and medical VQA (Section~\ref{sec:flipbench}; Appendix~\ref{app:variation-details}).\par Across 18 settings, \method improves the harmonic mean of these metrics by 2.8 points on average and up to 6.3 points over the strongest baseline in each setting. Ablation studies show that subspace estimation from contrastive pairs and margin-based gating both improve the balance between recovery and preservation. We will publicly release the complete benchmark, a reproducible construction and evaluation framework for additional models and perturbation settings, and all experimental code.}

Our contributions are threefold:
\begin{itemize}[leftmargin=1.3em,itemsep=2pt,topsep=4pt]
\item {We analyze \emph{answer flips} under subtle real-world visual variations, examining the associated activation shifts and decoding divergence.}
\item {We propose \method, a training-free inference-time method that estimates a low-rank flip-inducing activation subspace and uses margin-based gating to selectively steer hidden states during decoding.}
\item {We introduce \bench, a benchmark framework for model-specific evaluation of prediction recovery and preservation. It covers scientific reasoning, robot-scene understanding, and medical VQA under subtle visual variations, including lens contamination and medical image window-level changes.}
\end{itemize}

\section{Related Work}
\label{sec:related}

\paragraph{Robustness to real-world visual perturbations.}
Prior work studies natural corruptions, acquisition shifts, and semantics-preserving transformations \citep{imagenetc2019,medmnistc2024}. Recent benchmarks evaluate task accuracy under image corruptions \citep{mllmic2025} or semantic consistency between generated captions \citep{mmcbench2024}. Accuracy can reward prediction changes that happen to correct an original error. On a fixed evaluation set, high consistency may simply reflect variations that barely affect a given model. Evaluating recovery and preservation separately therefore requires identifying which predictions change and which remain stable for that model.

\paragraph{Jailbreaking and safety steering.}
Refusal can be mediated by a single activation direction \citep{refusal2024}, while multimodal defenses use activation differences associated with adversarial visual features to suppress harmful responses \citep{astra2025}. Such methods target an identifiable behavior.

\paragraph{Hallucination mitigation.}
Decoding methods improve visual grounding through visual contrast, token dependencies, or uncertainty \citep{vcd2024,opera2024,lead2026}; activation-based methods steer representations toward grounded behavior \citep{vti2024}.

\section{\method: Selective Steering for Answer-Flip Recovery}
\label{sec:method}
{\label{sec:characterization}  {\method estimates a low-rank flip-inducing activation subspace from activation differences between original inputs and their answer-flipping variants. During inference on new inputs, a margin-based gate selects decoding steps at which to attenuate the part of the hidden state that lies in this subspace (Figure~\ref{fig:method-overview}).} {Subspace estimation is performed offline using a set of contrastive pairs}; inference reuses the resulting basis {without gradient updates or auxiliary model training}.}

\subsection{Problem setup}
{\label{sec:problem} We study answer instability under mild visual changes that leave images nearly identical, such as exposure shifts, blur, and JPEG recompression. The objective is to recover the model's original answer when a perturbation changes it, while preserving answers on unaffected inputs. Let $f$ be a VLM with original answer $y=f(I,T)$ for image $I$ and prompt $T$. Let $\mathcal{P}$ denote the sampling distribution over transformations in a predefined perturbation family. For $\tau\sim\mathcal{P}$, let $\tilde I=\tau(I)$. A perturbation is \emph{flip-inducing} when}
\begin{equation}
\operatorname{norm}(f(\tilde I,T)) \neq \operatorname{norm}(f(I,T)),
\label{eq:flip}
\end{equation}
{where $\operatorname{norm}(\cdot)$ maps the model output to the task-specific answer space. Flip labels are model-specific because they reference the model's clean prediction. {Calibration is performed separately for each model, dataset, and known perturbation family; the resulting intervention is reused on new inputs from that setting without a clean reference or advance knowledge of their flip status.}}

\subsection{Estimating the flip-inducing subspace}
\label{sec:lowrank}
{\label{sec:subspace}  For each model and perturbation setting, let $\mathcal{D}_{\mathrm{flip}}=\{(I_i,\tilde I_i,T_i)\}_{i=1}^{n}$ {be a set of contrastive pairs} satisfying Eq.~\ref{eq:flip}, with at most one pair per original input and $n=|\mathcal{D}_{\mathrm{flip}}|$. This selection isolates answer-changing perturbations within families that can also preserve predictions.}

\paragraph{Paired activation differences.}
Let $h_\ell(I,T)\in\mathbb{R}^d$ be the hidden state at layer $\ell$ for the final prompt token after multimodal prefill. We extract this state from unedited passes on the clean and flipped inputs and compute
\begin{equation}
\delta_i=h_\ell(\tilde I_i,T_i)-h_\ell(I_i,T_i),\qquad \tilde I_i=\tau_i(I_i).
\label{eq:delta}
\end{equation}
Thus, $\delta_i$ measures the paired activation shift at the final prompt-token position. Estimating directions from contrastive examples and reusing them during generation follows prior activation-steering work \citep{refusal2024,rimsky2024caa,vti2024}. We use the final prompt token as a common comparison point before generated continuations diverge. We stack the differences from $n$ pairs into $\Delta=[\delta_1,\ldots,\delta_n]^\top\in\mathbb{R}^{n\times d}$.

\paragraph{Subspace estimation.}
We approximate these activation differences with a $k$-dimensional subspace. An orthonormal basis $U$ that minimizes the reconstruction error is obtained by
\begin{equation}
\min_{U^\top U=I_k}\|\Delta-\Delta UU^\top\|_F^2,
\qquad U=V_{:,1:k}\in\mathbb{R}^{d\times k},
\label{eq:subspace}
\end{equation}
{where $V$ contains the right singular vectors of $\Delta=P\Sigma V^\top$, ordered by decreasing singular value. Thus, $U$ retains $k$ singular directions, with $1\leq k\leq\operatorname{rank}(\Delta)\leq\min(n,d)$. We use uncentered SVD so that the mean difference, which may contain a shared clean-to-flip shift, contributes to the estimated subspace.   {The basis is computed once from {these contrastive pairs} and fixed during inference. {We call $\mathcal{S}_{\mathrm{flip}}=\operatorname{span}(U)\subseteq\mathbb{R}^d$ the estimated flip-inducing subspace. It summarizes activation shifts under answer-flipping perturbations and may also include other perturbation-related variation.}}}
\paragraph{Spectral concentration.}
{We measure the fraction of squared activation differences captured by the top $k$ directions as $E(k)=\sum_{j=1}^{k}\sigma_j^2/\sum_j\sigma_j^2$, where $\sigma_j$ are the singular values. For $255$ Qwen3-VL-8B/Robo2VLM lens-smudge pairs ($d=4096$), Figure~\ref{fig:energy-concentration} compares layers $12$ and $36$ with a Gaussian null; layer $4$ is shown in Appendix~\ref{app:energy-details} (Figure~\ref{fig:additional-evidence}a). The leading direction alone captures $15.6\%$ to $21.2\%$ of the energy, compared with $0.6\%$ for the null, and energy accumulates rapidly at small ranks. In this setting, the spectrum suggests that a compact subspace can capture a substantial fraction of the observed activation changes. Paired differences may also contain image- and prompt-specific variation, so retaining more directions need not improve steering on new inputs. We evaluate the effect of rank on steering performance in Section~\ref{sec:ablations}.}

\subsection{Margin-gated steering}
\label{sec:gating}
At inference, we use the estimated basis $U$ to attenuate the hidden-state component within the subspace. We apply this update selectively, using the next-token margin to determine when to intervene.

\begin{figure}[!t]
\centering
\begin{minipage}[t]{0.49\linewidth}
\centering
\includegraphics[width=\linewidth]{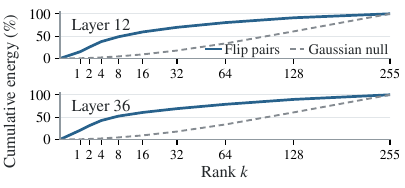}
\caption{\textbf{Low-rank concentration.} Leading directions capture more energy in comparison with a Gaussian null.}
\label{fig:energy-concentration}
\end{minipage}\hfill
\begin{minipage}[t]{0.49\linewidth}
\centering
\includegraphics[width=\linewidth]{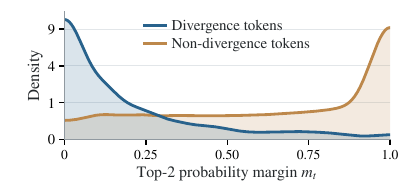}
\caption{\textbf{Next-token margins.} Divergence tokens tend to have lower margins in Qwen3-VL-8B.}
\label{fig:margin-divergence}
\end{minipage}
\end{figure}

\paragraph{Hidden-state steering.}
Let $h\in\mathbb{R}^d$ be the hidden state of the current token at the selected layer $\ell$. We steer this state by attenuating its projection onto the estimated subspace:
\begin{equation}
h'=h-\alpha UU^\top h
    =(I-UU^\top)h+(1-\alpha)UU^\top h.
\label{eq:ablation}
\end{equation}
The update scales the projection $UU^\top h$ by $1-\alpha$, leaving its orthogonal component unchanged. Thus, $\alpha=1$ removes the projection, while $0<\alpha<1$ partially attenuates it. At each step, an unedited forward pass first determines the margin and gate state. When the gate opens, we apply the hidden-state update and recompute the next-token distribution; otherwise, we retain the distribution from the unedited pass. We edit only decoding states; prompt prefill is unchanged.
During inference, steering uses only the current hidden state and stored basis, without a corresponding clean image, activation, or answer. Section~\ref{sec:setup-metrics} and Appendix~\ref{app:steering-implementation} detail the experimental settings and validation-based selection of $\ell$ and $\alpha$.
\paragraph{Next-token margin.}
Applying this update at every decoding step can also alter stable predictions. We therefore use the model's next-token uncertainty to select when to intervene.
{At decoding step $t$, we first run an unedited forward pass conditioned on the tokens generated so far. Let $p_t^{(1)}$ and $p_t^{(2)}$ be the two largest next-token probabilities. Their difference, $m_t=p_t^{(1)}-p_t^{(2)}$, is small when the model assigns similar probabilities to competing tokens.  To test its association with \emph{answer flips}, we analyze stored unedited generation traces. For flipped cases, we label differing next-token predictions under a shared prefix as \emph{divergence tokens}, realigning the prefix after each mismatch; matching predictions are \emph{non-divergence tokens}. Across $7$ Gemma and $8$ Qwen settings ($2{,}775$ divergence and $88{,}071$ non-divergence tokens), divergence tokens have lower median margins in every setting. Pooled over tokens, $93.0\%$ of divergence tokens have $m_t<0.3$, compared with $11.7\%$ of non-divergence tokens (Figure~\ref{fig:margin-divergence} for Qwen; Appendix~\ref{app:margin-details}, Figure~\ref{fig:additional-evidence}b for Gemma). This association motivates using a low margin to trigger steering.  The original input is used only to label tokens for this analysis; the gate does not require it at inference.}

\paragraph{Margin gate.}
With $\tau_{\mathrm{o}}<\tau_{\mathrm{c}}$, the stateful gate opens when $m_t<\tau_{\mathrm{o}}$, closes when $m_t>\tau_{\mathrm{c}}$, and retains its previous state otherwise. Gate decisions always use the unedited probabilities, so they do not depend on confidence introduced by the current edit. When the gate is open, we apply Eq.~\ref{eq:ablation}; otherwise, we use the original next-token distribution.
\section{\bench}
\label{sec:flipbench}

\bench is a benchmark framework for model-specific evaluation of prediction recovery and preservation under subtle visual variations. For a chosen model, dataset, and variation setting, it groups variants of each image--instruction pair according to whether they change the model's original prediction. Together with the original inputs, these flip and non-flip groups separately measure recovery of original predictions and preservation of stable ones, evaluating robustness rather than improvements in task accuracy (Section~\ref{sec:setup-metrics}). We instantiate the framework across 18 settings spanning scientific reasoning, robot-scene understanding, and medical VQA, with subtle visual variations common in each domain's real-world settings (Figure~\ref{fig:variations}).

\label{sec:flipbench-scope}
\label{sec:flipbench-build}
\label{sec:flipbench-stats}

\begingroup
\makeatletter
\renewcommand{\paragraph}{\@startsection{paragraph}{4}{\z@}{4pt}{-1em}{\normalfont\normalsize\bfseries}}
\makeatother
\paragraph{Science reasoning.}
We use the physics subset of M3CoT \citep{m3cot2024}, which requires multi-step reasoning over scientific diagrams and scenes. We consider exposure shift, white-balance shift, and double JPEG compression, covering changes in illumination, camera color processing, and image recompression during capture and transmission \citep{imagenetc2019,afifi2019whitebalance}.

\paragraph{Robot scenes.}
We use Robo2VLM \citep{robo2vlm2025}, which covers spatial relations, robot state, and action outcomes in manipulation scenes. We consider exposure shift, lens contamination, and motion blur, reflecting lighting changes, contamination of robot cameras, and camera or platform motion \citep{imagenetc2019,clp2026}.

\paragraph{Medical VQA.}
We use GMAI-MMBench \citep{gmai2024}, which spans clinical VQA tasks, perceptual granularities, and imaging modalities. We consider window-level variation, modality-specific defocus blur, and small detector rotation, representing variation in image display, optical or detector focus, and acquisition geometry \citep{bushberg2020essential,goodman2005fourier,mannan2016dfd}.
\par\endgroup

{Table~\ref{tab:benchmark-composition} lists parent-image counts by split and representative tasks or modalities.} GMAI-MMBench splits are stratified by clinical task and perceptual granularity, preserving these distributions and modality proportions. Pair search requires $12.8\pm6.4$ VLM calls on average, including unsuccessful searches (Table~\ref{tab:construction-cost}). Appendices~\ref{app:benchmark-construction} through~\ref{app:split-details} detail construction and filtering, and Appendix~\ref{app:medical} covers medical preprocessing. We will release the constructed evaluation sets and the framework for reproducible variant generation, model-specific flip labeling, and recovery--preservation evaluation.

\begin{table}[H]
\centering
\caption{Composition of \bench across three domains.}
\label{tab:benchmark-composition}
\small
\setlength{\tabcolsep}{3pt}
\begin{tabular*}{\linewidth}{@{\extracolsep{\fill}}>{\raggedright\arraybackslash}p{0.18\linewidth}>{\raggedright\arraybackslash}p{0.18\linewidth}rrr>{\raggedright\arraybackslash}p{0.40\linewidth}@{}}
\toprule
Domain & Source & Train & Val & Test & Coverage \\
\midrule
Science reasoning & M3CoT & 1,582 & 221 & 450 & Forces \& motion, energy, heat, magnetism \\
Robot scenes & Robo2VLM & 1,500 & 200 & 567 & Robot state, reachability, depth, next action \\
Medical VQA & GMAI-MMBench & 300 & 200 & 200 & X-ray, CT, endoscopy, fundus \\
\bottomrule
\end{tabular*}
\end{table}

\section{Experiments}
\label{sec:experiments}

We evaluate recovery, preservation, transfer, and computational cost on \bench using two open-weight VLMs and inference-time baselines.

\subsection{Experimental Setup}
\label{sec:setup-metrics}
\begingroup
\setlength{\parskip}{4pt}

\noindent\textbf{Models.}
Since \method requires access to intermediate hidden states, we evaluate it on two open-weight VLMs, Gemma-3-12B-IT \citep{gemma32025} and Qwen3-VL-8B-Instruct \citep{qwen3vl2025}.
\par

{\noindent\textbf{Baselines.} We compare against the unmodified model (Base), two decoding-time methods, VCD \citep{vcd2024} and LEAD \citep{lead2026}, and the activation-steering method VTI \citep{vti2024}. Implementation details appear in Appendix~\ref{app:baseline-details}.}

\noindent\textbf{Metrics.} We evaluate each model on its own original, flip, and non-flip groups: preservation of original predictions on clean ($\mathrm{C}$) and non-flip ($\mathrm{N}$) inputs, and recovery of those predictions on flipped inputs ($\mathrm{F}$). Our primary metric is their harmonic mean,
\begin{equation}
\hrec=\left[\frac{1}{3}\left(\frac{1}{\mathrm{C}}+\frac{1}{\mathrm{N}}+\frac{1}{\mathrm{F}}\right)\right]^{-1},
\label{eq:harmonic-mean}
\end{equation}
{with $\hrec=0$ if any component is zero. All three metrics use the model's original prediction as the reference, measuring robustness to visual variations rather than task accuracy. Recovery measures restoration of that prediction, including when it is incorrect. Reporting $\mathrm{F}$ separately from $\mathrm{C}$ and $\mathrm{N}$ prevents stable cases from masking poor recovery.}

\noindent\textbf{Implementation details.}
\label{sec:tuning}
{We use rank $k=8$ and shared gate thresholds $(\tau_{\mathrm{o}},\tau_{\mathrm{c}})=(0.3,0.7)$ across all settings. For each model and perturbation setting, we estimate the flip subspace from the training split and select the steering layer $\ell$ and strength $\alpha$ on the validation split. The test split is used only for final evaluation. {We use deterministic decoding where available for reproducibility.}  Implementation and selection details appear in Appendix~\ref{app:experimental-details}.}

\par
\endgroup

\subsection{Main Results}
\label{sec:main-results}
{\method achieves the highest $\hrec$ in all 18 settings, exceeding the strongest baseline by $2.8$ points on average (Tables~\ref{tab:main-qwen} and~\ref{tab:main-gemma}). Mean gains are $2.5$, $2.5$, and $3.4$ points for science, robotics, and medical VQA, and $2.4$ and $3.2$ points for Qwen3-VL-8B and Gemma-3-12B, respectively. }
{These gains reflect a better balance between recovering original predictions and preserving stable ones, rather than higher task accuracy. On Gemma/Robo2VLM exposure, \method slightly reduces recovery relative to LEAD ($55.0$ versus $56.4$) but improves $\mathrm{C}/\mathrm{N}$ from $84.5/85.3$ to $88.2/87.8$, raising $H$ from $72.7$ to $73.4$. On Qwen/M3CoT JPEG, it instead raises recovery from the strongest baseline's $56.1$ to $66.3$ while keeping $\mathrm{C}/\mathrm{N}$ above $87\%$, improving $H$ by $5.1$ points.  }

\begin{figure}[t]
\centering
\includegraphics[width=\linewidth]{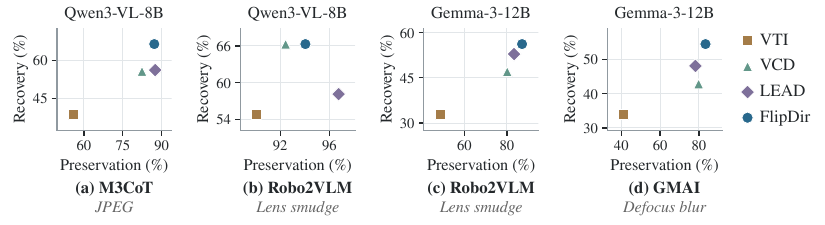}
\caption{ {\textbf{Recovery--preservation trade-offs in selected settings.} Each point is a reported configuration, not a tuning sweep.  All 18 settings, including Base, appear in Appendix~\ref{app:tradeoffs}.}}
\label{fig:tradeoff}
\end{figure}

{ {Figure~\ref{fig:tradeoff} compares flip recovery ($\mathrm{F}$) with average clean and non-flip preservation, $P=(\mathrm{C}+\mathrm{N})/2$, in four settings.} Across all 18 settings, \method is Pareto-nondominated among the reported operating points, with some baselines also on the frontier (Appendix~\ref{app:tradeoffs}).}

\subsection{Transfer Across Datasets and Perturbation Combinations}
\label{sec:transfer}

\begin{figure}[!htbp]
\centering
\captionsetup{font=footnotesize}
\begin{minipage}[t]{0.62\linewidth}
\vspace{0pt}
\centering
\includegraphics[width=\linewidth]{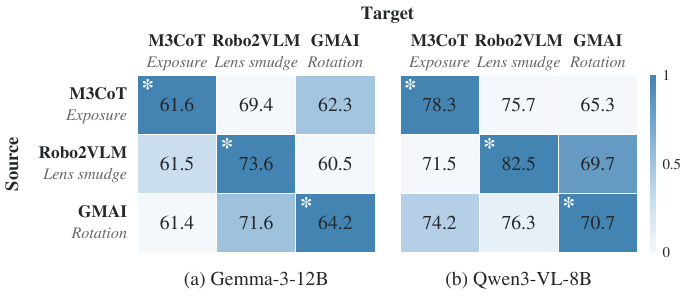}
\caption{\textbf{Flip-subspace transfer} ($H$, \%). Column-wise min--max colors compare sources for each target. White asterisks mark matched source--target pairs.}
\label{fig:analysis2}
\end{minipage}\hfill
\begin{minipage}[t]{0.355\linewidth}
\vspace{0pt}
\centering
\includegraphics[width=\linewidth]{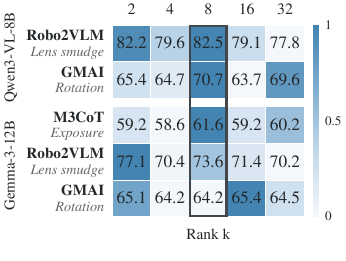}
\caption{\textbf{Rank sensitivity} ($H$, \%). Row-normalized colors; boxed $k=8$. GMAI denotes GMAI-MMBench.}
\label{fig:rank-sensitivity}
\end{minipage}
\end{figure}

{We transfer only $U$ across datasets and perturbation combinations, retaining target-specific layers and strengths (Figure~\ref{fig:analysis2}). Across the twelve off-diagonal pairs, $H$ decreases by $0.1$ to $6.8$ points relative to the target-matched subspace. The largest relative decrease occurs for Robo2VLM lens smudge $\rightarrow$ M3CoT exposure on Qwen ($71.5$ versus $78.3$, retaining $91\%$ of target-matched $H$). This probes shared subspace structure under target-specific calibration, rather than calibration-free deployment; dataset and perturbation effects are not isolated. Appendix~\ref{app:transfer-details} gives full results.}
\subsection{Ablation Studies}
\label{sec:ablations}

\paragraph{Representation difference.}
\label{sec:analysis}
\label{sec:components}
{Flip-inducing and non-flipping perturbations yield similar mean activation-shift norms ($15.2$ versus $15.6$) but differ in direction, suggesting that \emph{answer flips} are associated with the direction rather than the magnitude of the  representation change. On Robo2VLM exposure and GMAI window level, we re-estimate $U$ for Gemma-3-12B from three representation differences while holding all other components fixed (Table~\ref{tab:uvar}).  {\emph{Flip $-$ Clean} achieves the highest $H$ in both evaluations.  These results suggest that conditioning on \emph{answer flips} enriches perturbation-induced representation shifts for directions useful for recovery.}}

\begin{table}[!htbp]
\captionsetup{font=footnotesize}
\begin{minipage}[t]{0.40\linewidth}

\centering
\setlength{\tabcolsep}{2pt}
\footnotesize
\renewcommand{\arraystretch}{0.95}
\caption{Representation ablation ($H$, \%).}
\label{tab:uvar}
\begin{tabular*}{\linewidth}{@{\extracolsep{\fill}}lcc@{}}
\toprule
& \multicolumn{2}{c}{Gemma-3-12B} \\\cmidrule(lr){2-3}Representation & Robo2VLM & GMAI \\
\midrule
Flip $-$ Clean & \textbf{73.4} & \textbf{70.0} \\
Non-flip $-$ Clean & 61.5 & 67.9 \\
Flip $-$ Non-flip & 68.8 & 64.6 \\
\bottomrule
\end{tabular*}

\end{minipage}\hfill
\begin{minipage}[t]{0.58\linewidth}

\centering
\caption{Margin-gate ablation ($H$, \%); $\Delta=\mathrm{On}-\mathrm{Off}$.}
\label{tab:ablation-main}
\footnotesize
\setlength{\tabcolsep}{2pt}
\renewcommand{\arraystretch}{0.95}\begin{tabular*}{\linewidth}{@{\extracolsep{\fill}}lrrrrrr@{}}
\toprule
& \multicolumn{3}{c}{Gemma-3-12B} & \multicolumn{3}{c}{Qwen3-VL-8B} \\
\cmidrule(lr){2-4}\cmidrule(lr){5-7}
Setting & Off & On & $\Delta$ & Off & On & $\Delta$ \\
\midrule
M3CoT exp. & 54.8 & \textbf{61.6} & \textcolor{green!45!black}{+6.8} & 75.6 & \textbf{78.3} & \textcolor{green!45!black}{+2.7} \\
Robo2VLM lens & 68.8 & \textbf{73.6} & \textcolor{green!45!black}{+4.8} & 73.5 & \textbf{82.5} & \textcolor{green!45!black}{+9.0} \\
GMAI rot. & 58.3 & \textbf{64.2} & \textcolor{green!45!black}{+5.9} & 69.2 & \textbf{70.7} & \textcolor{green!45!black}{+1.5} \\
\bottomrule
\end{tabular*}

\end{minipage}
\end{table}

\newsavebox{\flipdirAblationFigBox}
\begin{lrbox}{\flipdirAblationFigBox}
\begin{minipage}[t]{0.32\linewidth}
\vspace{0pt}
\setlength{\parskip}{0pt}
\centering
\includegraphics[width=\linewidth]{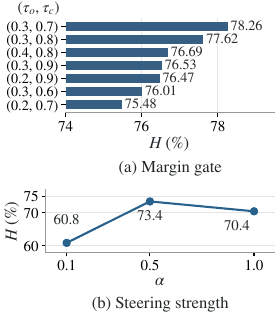}
\captionsetup{type=figure,font=footnotesize,skip=3pt}
\caption{\textbf{Gate and strength sensitivity.} Other components are fixed.}
\label{fig:analysis}
\par\vspace{0pt}
\end{minipage}
\end{lrbox}
\noindent
\begin{minipage}[t][\dimexpr\ht\flipdirAblationFigBox+\dp\flipdirAblationFigBox\relax][s]{0.655\linewidth}
\vspace{0pt}
\setlength{\parskip}{4pt}
\paragraph{Margin gate.}
Table~\ref{tab:ablation-main} compares margin-gated steering with steering at every decoding step, holding other components fixed within each setting. The gate improves $H$ in all six comparisons. Figure~\ref{fig:gate-gains} in Appendix~\ref{app:component-ablation} shows gains in C, N, F, and H; Figure~\ref{fig:analysis}(a) varies thresholds on Qwen3-VL-8B/M3CoT exposure, with the $H$ axis starting at $74\%$.

\par\vfill
\paragraph{Steering strength.}
{Figure~\ref{fig:analysis}(b) varies $\alpha$ on Gemma-3-12B/Robo2VLM exposure. Partial attenuation at $\alpha=0.5$ achieves higher $H$ than either weaker steering at $\alpha=0.1$ or full projection removal at $\alpha=1$.  Each sweep evaluates one setting.}

\par\vfill
\paragraph{Subspace rank.}
{Across the five settings in Figure~\ref{fig:rank-sensitivity}, increasing $k$ beyond $8$ yields no consistent gain, although the best rank varies by setting.  Additional ablations appear in Appendix~\ref{app:ablation}.}
\par\vspace{0pt}
\end{minipage}\hfill
\usebox{\flipdirAblationFigBox}
\par

\paragraph{Computational efficiency.}
\label{sec:budget}
{\label{sec:cost} \method extracts paired activations and computes $U$ by SVD once for reuse across inputs. At each decoding step, an unedited forward supplies the margin for gating. An additional forward pass applies steering only when the gate is open. With lower-layer activations reused, estimated decoding FLOPs increase by $8\%$ on average at matched sequence lengths (Table~\ref{tab:compute}). Appendices~\ref{app:compute-accounting} and~\ref{app:runtime} detail offline costs, estimation assumptions, and measured runtimes.}
\begin{table}[!htbp]
\centering
\caption{Offline preparation and mean estimated decoding FLOPs relative to Base.}
\label{tab:compute}
\small
\setlength{\tabcolsep}{7pt}
\begin{tabular}{@{}lccccc@{}}
\toprule
 & Base & LEAD & VTI & VCD & \method \\
\midrule
Offline directions & No & No & Yes & No & Yes \\
Decode FLOPs & $1.00\times$ & $\approx1.08\times$ & $\approx1.00\times$ & $\approx2.00\times$ & $\approx1.08\times$ \\
\bottomrule
\end{tabular}
\end{table}
\section{Conclusion}
{\label{sec:conclusion}We study \emph{answer flips} under subtle visual changes common in real-world image capture, processing, and transmission. Observations of activation spectra and token margins motivate \method, which selectively attenuates a learned flip subspace during decoding without weight updates or a clean reference. \bench evaluates robustness relative to each model's original predictions, separating recovery from preservation across scientific reasoning, robot-scene understanding, and medical VQA. Across 18 settings, \method improves the harmonic mean of preservation and recovery by $2.8$ points on average over the strongest baseline.}

\paragraph{Limitations and future work.}\label{sec:limitations} \method does not yet identify a single subspace that generalizes across perturbation types and supports steering throughout decoding without disrupting stable predictions. Future work will investigate whether such a shared subspace can remove the need for setting-specific calibration and margin gating.

\newpage
\subsection*{AI use statement}
We used LLMs or other AI tools to aid or polish writing, for research ideation or execution, to draft sections of the paper, and for the implementation of experimental code. The authors guided and reviewed all AI-assisted work and take full responsibility for the final manuscript and associated artifacts.

\subsection*{Reproducibility statement}
We will publicly release the complete \bench dataset, the code used to construct it, and all experimental code, including implementations of \method and the evaluated baselines, evaluation scripts, and analysis and figure-generation scripts. Section~\ref{sec:method} describes the method, and Section~\ref{sec:flipbench} and Appendix~\ref{app:datasets} document benchmark construction, perturbation settings, and data splits. Appendix~\ref{app:frontier-analysis} specifies the evaluation pools and budget-matching protocol. Appendix~\ref{app:experimental-details} provides implementation details, hyperparameter selection, baseline adaptations, and inference settings. The release will also include versioned configurations, random seeds and provenance metadata, and the extracted flip subspaces, as described in Appendix~\ref{app:repro}. Computational cost accounting is provided in Appendix~\ref{app:compute-accounting}.

\bibliography{flipdir}
\bibliographystyle{iclr2027_conference}

\appendix
\clearpage
\makeatletter
\setlength{\@fptop}{0pt} %
\makeatother

\FloatBarrier
\section{Benchmark Construction and Data}
\label{app:datasets}

\subsection{Benchmark construction}
\label{app:benchmark-construction}Algorithm~\ref{alg:builder} instantiates the \bench framework for a specified dataset, model, and variation family. Applying it to another model requires obtaining that model's original predictions and flip labels, rather than reusing labels from another model. We first obtain a greedy clean-image answer and exclude parents whose answer cannot be normalized to the option space. Perturbations are then sampled from the fixed ranges in Table~\ref{tab:variations} using a per-parent seed chain and evaluated with the same prompt. Each child is labeled relative to its parent's clean prediction using Eq.~\ref{eq:flip}. Full-budget runs evaluate all $B$ draws; pair-existence runs may stop once both a flip and a non-flip child are found. The released field \texttt{stable} denotes non-flip children. Seeds, parameters, prompts, outputs, and parent identifiers are retained for reconstruction.

\begin{algorithm}[H]
\caption{Answer-flip benchmark construction (full-budget variant)}
\label{alg:builder}
\begin{algorithmic}[1]
\Require Base dataset $\mathcal{D}$, VLM $f$, perturbation family $\mathcal{P}$, budget $B$
\ForAll{$(I, T, y^\star) \in \mathcal{D}$}
  \State $y_{\mathrm{clean}} \gets f(I,T)$
  \If{$y_{\mathrm{clean}}$ cannot be normalized}
    \State \textbf{continue}
  \EndIf
  \For{$b = 1$ to $B$}
    \State Sample $\tau_b \sim \mathcal{P}$ using a deterministic seed chain
    \State $\tilde I_b \gets \tau_b(I)$; \quad $\tilde y_b \gets f(\tilde I_b,T)$
    \State $z_b \gets \mathbb{1}[\operatorname{norm}(\tilde y_b) \neq \operatorname{norm}(y_{\mathrm{clean}})]$
    \State Store image, seed, variation parameters, raw output, normalized answer, and label $z_b$
  \EndFor
\EndFor
\end{algorithmic}
\end{algorithm}

\paragraph{Pair-search cost.}
Pair search stops when both child types are found, after $15$ draws without a flip, or at the $50$-draw cap. This stopping rule differs from the full-budget procedure in Algorithm~\ref{alg:builder}.

Table~\ref{tab:construction-cost} reports the number of VLM calls under this stopping rule, including the clean call and subsequent perturbed-image calls. Each search is defined by a parent, a perturbation family, and a backbone; the same source parent may therefore contribute to multiple searches. Across $27{,}478$ searches, construction uses $352{,}970$ calls: $87{,}384$ for searches that obtain both child types and $265{,}586$ for those that do not. Counts include all calls made within each search, not only the retained children.

\begin{table}[!htbp]
\centering
\caption{\textbf{VLM calls for benchmark construction.} Calls per search are reported as mean $\pm$ standard deviation. Pair found denotes obtaining both a flip and a non-flip child.}
\label{tab:construction-cost}
\small
\setlength{\tabcolsep}{3pt}
\begin{tabular}{l rcr rcr r}
\toprule
& \multicolumn{3}{c}{Pair found} & \multicolumn{3}{c}{Pair not found} & \\
\cmidrule(lr){2-4}\cmidrule(lr){5-7}
Domain & Searches & Calls/search & Calls & Searches & Calls/search & Calls & All calls \\
\midrule
\multicolumn{8}{l}{\textit{Gemma-3-12B}} \\
M3CoT & 3,817 & $6.25\pm5.08$ & 23,858 & 2,276 & $16.82\pm5.28$ & 38,271 & 62,129 \\
Robo2VLM & 1,485 & $6.99\pm5.09$ & 10,387 & 4,716 & $16.10\pm1.90$ & 75,946 & 86,333 \\
GMAI-MMBench & 781 & $6.43\pm5.00$ & 5,021 & 719 & $17.02\pm5.90$ & 12,239 & 17,260 \\
\midrule
\multicolumn{8}{l}{\textit{Qwen3-VL-8B}} \\
M3CoT & 2,971 & $9.71\pm7.14$ & 28,847 & 3,062 & $16.45\pm3.93$ & 50,357 & 79,204 \\
Robo2VLM & 1,312 & $10.02\pm7.46$ & 13,148 & 4,863 & $16.11\pm1.94$ & 78,333 & 91,481 \\
GMAI-MMBench & 876 & $6.99\pm5.22$ & 6,123 & 600 & $17.40\pm6.86$ & 10,440 & 16,563 \\
\midrule
All & 11,242 & $7.77\pm6.21$ & 87,384 & 16,236 & $16.36\pm3.52$ & 265,586 & \textbf{352,970} \\
\bottomrule
\end{tabular}
\end{table}

\subsection{Perturbation definitions and ranges}
\label{app:variation-details}
\begin{table}[!htbp]
\centering
\caption{\textbf{Perturbation ranges.} Each instance uses a deterministic seed chain. Appendix~\ref{app:medical} specifies modality-dependent medical operators.}
\label{tab:variations}
\small
\begin{tabular}{lll}
\toprule
Domain & Variation & Sampling band \\
\midrule
M3CoT & Exposure shift & $|\text{EV}| \in [0.45, 0.65]$ \\
M3CoT & White-balance shift & $1000$ to $3000$ K \\
M3CoT & Double JPEG & $q_1 \in [85, 95]$, $q_2 \in [72, 85]$, $q_2 \neq q_1$ \\
Robo2VLM & Exposure shift & $|\text{EV}| \in [0.45, 0.65]$ \\
Robo2VLM & Lens smudge & wipe smears, droplets, dust; opacity ${\leq}0.6$, blur $\sigma=4.5$ \\
Robo2VLM & Motion blur & inertial trajectory, intensity $0.35$ \\
GMAI-MMBench & Window-level shift & contrast and gamma in $[0.80, 1.25]$ \\
GMAI-MMBench & Defocus blur & modality-gated PSF (disk or Gaussian) \\
GMAI-MMBench & Detector rotation & $\pm 1.5^{\circ}$ to $4^{\circ}$, reflect-pad and center crop \\
\bottomrule
\end{tabular}
\end{table}

\paragraph{Perturbation operators.}
Exposure adjusts image intensity in EV units; white balance applies a global illuminant shift \citep{afifi2019whitebalance}; double JPEG recompresses the image twice at distinct quality levels. Lens smudge composites one to two Bezier wipe smears (opacity $0.28$ to $0.45$), two to four droplets ($0.30$ to $0.50$), and two to six dust specks, with local blur ($\sigma=4.5$), $22\%$ haze, and an opacity cap of $0.6$ \citep{clp2026}. Motion blur uses an inertial camera-shake trajectory at intensity $0.35$. Medical operators are specified in Appendix~\ref{app:medical}. Reported $\Delta E$ is the mean per-pixel CIE76 difference in CIELAB space.

\subsection{Dataset composition and task coverage}
\label{app:composition-details}
Table~\ref{tab:benchmark-composition} summarizes the parent-image pools used in the main benchmark. M3CoT contributes eight physics topics covering scientific diagrams and multi-step conceptual reasoning. Robo2VLM contributes eight question types spanning robot state, reachability, spatial correspondence, depth, action outcomes, and grasping. GMAI-MMBench is organized by clinical VQA task, perceptual granularity, and imaging modality, with both lens-based and detector-based images represented. {Parents in the training split serve as offline calibration data to estimate $U$, validation parents to select the layer $\ell$ and strength $\alpha$, and test parents for final evaluation. These pools are disjoint. Validation selection uses parents with both a flip and a non-flip child; its effective size therefore depends on the setting (Appendix~\ref{app:selection-details}).}

\subsection{Split construction and filtering}
\label{app:split-details}

For M3CoT and Robo2VLM, we retain the domain and task definitions of the source datasets and apply model-specific usability filtering only after obtaining the clean prediction. Parents whose clean answer cannot be normalized to the task answer space are excluded, which is why the usable evaluation counts can differ slightly across models (Table~\ref{tab:app-sizes}).

For GMAI-MMBench, we filter the public validation release and partition it with fixed-seed largest-remainder allocation over joint clinical-task and perceptual-granularity strata. Task, granularity, and modality marginals differ by about two percentage points or less between the resulting train and test pools. Appendix~\ref{app:medical} provides the modality-specific filtering and image-processing details.

\subsection{Medical data and preprocessing}
\label{app:benchmark-design}
\label{app:medical}
We remove GMAI-MMBench questions with fewer than four options or answerable by a single text-only greedy query, leaving $2{,}397$ candidates. Lens-based modalities (endoscopy, fundus, dermoscopy, and microscopy) use a disk point-spread function for defocus \citep{goodman2005fourier,mannan2016dfd}; X-ray and CT use a Gaussian kernel \citep{bushberg2020essential}. MRI is outside this isotropic-blur protocol. Window-level shift composes contrast and gamma changes in $[0.80,1.25]$. Rotation uses reflect padding and center cropping to model small in-plane alignment changes.

Images are downscaled, when needed, to a maximum long side of $1280$ pixels for Qwen3-VL and $896$ for Gemma-3, following the VLMEvalKit convention. Perturbations are applied after resizing, with blur radius defined relative to the short side. Table~\ref{tab:app-sizes} reports the model-specific subspace and evaluation pools.

\FloatBarrier
\Needspace{25\baselineskip}
\section{Evaluation Protocol and Answer-Flip Statistics}
\label{app:frontier-analysis}

\subsection{Evaluation pool sizes}
\label{app:pool-sizes}
Table~\ref{tab:app-sizes} lists the calibration pairs and the actual evaluation denominators for each setting.
\begin{table}[H]
\centering
\caption{\textbf{Subspace and evaluation pool sizes.} $U$ pairs counts the pairs used to estimate the subspace, at most one per parent. Eval C/N/F lists the actual clean, non-flip, and flip denominators used for recovery metrics. Parent pools are given in Table~\ref{tab:benchmark-composition}.}
\label{tab:app-sizes}
\small
\begin{tabular}{l cc cc}
\toprule
& \multicolumn{2}{c}{Qwen3-VL-8B} & \multicolumn{2}{c}{Gemma-3-12B} \\
\cmidrule(lr){2-3}\cmidrule(lr){4-5}
Setting & $U$ pairs & Eval C/N/F & $U$ pairs & Eval C/N/F \\
\midrule
M3CoT exposure          &  737 & 429/428/183 &  989 & 449/447/278 \\
M3CoT white balance     &  723 & 429/428/190 &  977 & 449/448/272 \\
M3CoT JPEG              &  759 & 429/428/187 & 1039 & 449/444/298 \\
Robo2VLM exposure       &  334 & 560/558/107 &  316 & 567/566/149 \\
Robo2VLM lens smudge    &  255 & 560/554/86 &  319 & 567/566/153 \\
Robo2VLM motion blur    &  287 & 560/558/94 &  375 & 567/564/187 \\
GMAI window level       &  183 & 192/190/112 &  185 & 200/199/102 \\
GMAI defocus blur       &  146 & 192/189/85 &  120 & 200/191/77 \\
GMAI detector rotation  &  185 & 192/192/110 &  192 & 200/197/126 \\
\bottomrule
\end{tabular}
\end{table}

\subsection{Budget-matched cross-model evaluation}
\label{app:kmatch}

Cross-model comparisons use a common budget of $B=15$ perturbation draws per parent. For open-weight models, recorded children are matched to reconstructed draws with a tolerance of two pixel-intensity levels, and only the first 15 draws are counted. API models use this budget directly. These rates use a separate parent pool from the test-split percentages in Table~\ref{tab:fliprates}.

Table~\ref{tab:prevalence} reports clean accuracy and the \emph{answer-flip rate} under this protocol. All four models exhibit flips, including the two closed models. In this comparison, flip rate decreases as clean accuracy increases, but not with parameter count. On flipped inputs, the two open-weight models achieve only about $30\%$ accuracy, showing that most flips are errors rather than harmless rephrasings. Flips also occur on many of the same inputs across models. A question that flips Qwen3-VL-8B is more likely to flip Claude Opus 4.8 (odds ratio $5.2$, $95\%$ CI $[3.1, 8.8]$), and $31$ of GPT-5.5's $41$ flips occur on parents that also flip Qwen3-VL-8B. These results indicate overlap in the inputs on which different models exhibit answer instability.

\subsection{Answer-flip rates across all evaluation settings}
\label{app:fliprates}
Table~\ref{tab:fliprates} reports answer-flip percentages in the evaluation pools, using the clean and flip counts in Table~\ref{tab:app-sizes}. Cross-model comparisons use the separate matched-budget protocol above.

\begin{table}[!htbp]
\centering
\caption{\textbf{Answer flips in the evaluation pools.} Percentages are $100n_F/n_C$, using the flip and clean counts in Table~\ref{tab:app-sizes}; $n=n_C$.}
\label{tab:fliprates}
\small
\begin{tabular}{lcc}
\toprule
Setting & Qwen3-VL-8B & Gemma-3-12B \\
\midrule
M3CoT exposure & 42.7\% (n=429) & 61.9\% (n=449) \\
M3CoT white balance & 44.3\% (n=429) & 60.6\% (n=449) \\
M3CoT JPEG & 43.6\% (n=429) & 66.4\% (n=449) \\
Robo2VLM exposure & 19.1\% (n=560) & 26.3\% (n=567) \\
Robo2VLM lens smudge & 15.4\% (n=560) & 27.0\% (n=567) \\
Robo2VLM motion blur & 16.8\% (n=560) & 33.0\% (n=567) \\
GMAI window level & 58.3\% (n=192) & 51.0\% (n=200) \\
GMAI defocus blur & 44.3\% (n=192) & 38.5\% (n=200) \\
GMAI detector rotation & 57.3\% (n=192) & 63.0\% (n=200) \\
\bottomrule
\end{tabular}
\end{table}

\FloatBarrier
\section{Implementation and Reproducibility}
\label{app:experimental-details}

\subsection{\method configuration}
\label{app:config}
We select the steering layer and strength separately for each model and perturbation setting. Table~\ref{tab:app-config} lists the selected values and corresponding main-table results. All settings share rank $k=8$ and gate thresholds $(\tau_o,\tau_c)=(0.3,0.7)$.

\begin{table}[!htbp]
\centering
\caption{\textbf{Selected configurations.} Layer $\ell$, strength $\alpha$, and main-table $\hrec$. All settings use $k=8$, gate thresholds $(0.3,0.7)$, and uncentered SVD with at most one flip pair per parent.}
\label{tab:app-config}
\small
\begin{tabular}{l ccc ccc}
\toprule
& \multicolumn{3}{c}{Gemma-3-12B} & \multicolumn{3}{c}{Qwen3-VL-8B} \\
\cmidrule(lr){2-4}\cmidrule(lr){5-7}
Setting & $\ell$ & $\alpha$ & $\hrec$ & $\ell$ & $\alpha$ & $\hrec$ \\
\midrule
M3CoT exposure       & 12 & 1.0 & 61.6 &  4 & 1.0 & 78.3 \\
M3CoT white balance  & 12 & 0.5 & 60.8 & 36 & 1.0 & 76.4 \\
M3CoT JPEG           & 12 & 1.0 & 63.0 & 36 & 1.0 & 78.9 \\
Robo2VLM exposure    & 12 & 0.5 & 73.4 & 36 & 1.0 & 82.4 \\
Robo2VLM lens smudge & 36 & 0.5 & 73.6 & 36 & 1.0 & 82.5 \\
Robo2VLM motion blur & 36 & 1.0 & 69.1 & 36 & 0.5 & 82.4 \\
GMAI window level    & 12 & 0.5 & 70.0 & 12 & 0.5 & 70.1 \\
GMAI defocus blur    & 36 & 1.0 & 70.9 & 36 & 1.0 & 63.7 \\
GMAI rotation        & 36 & 0.5 & 64.2 & 36 & 0.5 & 70.7 \\
\bottomrule
\end{tabular}
\end{table}

\subsection{Steering implementation}
\label{app:steering-implementation}
We extract the states used to estimate $U$ at the final prompt position in unedited prefill, but apply the resulting directions only during decoding at the selected layer. The projection can be evaluated as $U(U^\top h)$ using two matrix-vector products, requiring $O(dk)$ arithmetic without materializing a $d\times d$ projection matrix. At each decoding step, we first obtain the unedited distribution and update the gate from its top-two probability margin. An open gate triggers a second forward with the projection applied at layer $\ell$; a closed gate uses the first distribution without a steered forward.

\subsection{Hyperparameter selection}
\label{app:selection-details}
The six $(\ell,\alpha)$ candidates are scored on validation triplets containing a clean image, one flip child, and one non-flip child. Parents follow a fixed 40-shard ordering; the first $R$ shards are used to cover approximately 50 flip-bearing parents, with $R$ adjusted to the setting's flip density. Qwen3-VL defocus blur has 47 available validation flip pairs and approximately 24 selection triplets. {Validation parents are disjoint from subspace-calibration and test parents.} Varied images are stored as PNG at construction time and are regenerable exactly from the deterministic per-parent seed chain.

\subsection{Baseline adaptations}\label{app:baseline-details} {We use the recommended LEAD and VTI configurations without task-specific retuning. The adaptations below address backbone compatibility. In contrast, \method selects its method-specific intervention layer and strength on held-out validation data (Appendix~\ref{app:selection-details}); test data are used only for final evaluation.}

\paragraph{LEAD.}\label{app:lead}
We adapt the public LEAD implementation for Qwen2.5-VL \citep{lead2026} to each model's embedding and token conventions. For Gemma-3, direct embedding inputs retain the model's $\sqrt{d}$ scaling, termination recognizes its chat end token, and image-anchor lookup uses the model-specific image token. Table~\ref{tab:main-gemma} reports this adapted implementation.

\paragraph{VTI.}\label{app:vti}
We use the official VTI implementation. Qwen3-VL produces a variable visual-token grid, so we average VTI's visual direction over tokens within each layer and apply it at all positions. The textual direction is unchanged. Setting $\alpha=\beta=0$ reproduces the base model exactly.

\subsection{Deterministic inference and released artifacts}
\label{app:repro}
\label{app:deterministic-inference}
All open-weight experiments use greedy decoding, batch size one, fixed seeds, deterministic kernels, and TF32 disabled. Repeating one complete setting reproduced all $1{,}040$ generations exactly. Exact reproduction uses the saved flip subspaces: GPU re-extraction can introduce numerical differences, and subspaces agreeing to $10^{-6}$ can still change $H$ by several points at near-tied decoding steps.

\paragraph{Release contents.}\label{app:released-artifacts}\label{app:repro-checklist}
We will release model-specific benchmark instances, perturbation code and versioned configurations, inference and evaluation scripts, answer-normalization and flip-mining utilities, budget-matching tools, and extracted subspaces. Instance metadata includes the parent identifier, seed, perturbation parameters, prompt, raw output, and normalized answer.

\FloatBarrier
\section{Activation and Decoding Analyses}
\label{app:diagnostics}

\subsection{Activation energy concentration}
\label{app:energy-details}
For Qwen3-VL-8B on Robo2VLM lens smudge, we analyze $255$ clean-flip pairs ($d=4096$) at layers $4$, $12$, and $36$. The Gaussian null uses a matrix of the same size and the same uncentered SVD and energy calculation. With uncentered singular values $\sigma_{\ell,j}$, cumulative energy is $E_\ell(k)=\sum_{j=1}^{k}\sigma_{\ell,j}^{2}/\sum_j\sigma_{\ell,j}^{2}$. The top $8$ directions capture $48.6$ to $53.7\%$ of the energy, versus $4.7\%$ for the Gaussian null; the top $16$ capture $59.4$ to $65.7\%$, versus $9.2\%$.
Figures~\ref{fig:energy-concentration} and~\ref{fig:additional-evidence}a plot these cumulative energies.

The cumulative-energy plots use a square-root rank axis and a linear energy axis.

\subsection{Token margins at decoding divergence}
\label{app:margin-details}
\begin{figure}[!htbp]
\centering
\includegraphics[width=0.70\linewidth]{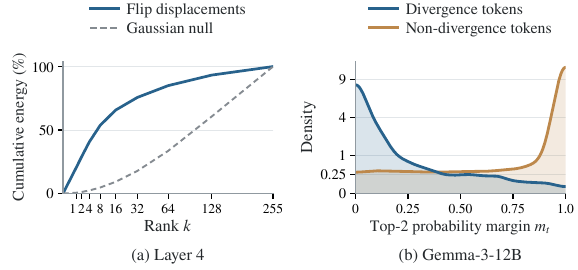}
\caption{\textbf{Additional activation and decoding analyses.} (a) Cumulative energy at layer $4$ for the same $255$ Qwen3-VL-8B/Robo2VLM lens-smudge pairs as Figure~\ref{fig:energy-concentration}. (b) Divergence and non-divergence token margins for Gemma-3-12B, complementing the Qwen results in Figure~\ref{fig:margin-divergence}.}
\label{fig:additional-evidence}
\end{figure}

Figure~\ref{fig:additional-evidence}(b) complements the Qwen3-VL-8B results in Figure~\ref{fig:margin-divergence} with Gemma-3-12B. We analyze stored unedited traces for the settings with available traces. At each step, original and perturbed inputs share the same decoding prefix. A divergence occurs when their next-token predictions differ; after each divergence, we advance both runs with the same token to realign the prefix before continuing. Margins are computed from the perturbed-input distribution. Within each backbone, divergence and non-divergence token distributions are normalized separately to unit area. We use Gaussian kernel density estimation with bandwidth $0.045$ and reflection at $0$ and $1$. The square-root density axis makes overlap visible; tick labels report density in the original units.

\subsection{Limits of static diagnostics}
\label{app:selection-proxies} {We examine whether simple diagnostics predict \emph{answer flips} or recovery. Subspace alignment has little correlation with realized recovery ($\rho\approx0$), and steering coordinates provide weak token-level discrimination (SNR $0.02$ to $0.07$; AUC $0.55$). A parent's flip probability is also nearly flat in its clean-run margin ($0.48$ to $0.60$). These results limit the use of alignment, steering coordinates, or clean-run confidence as standalone predictors.} {\method does not use $U$ as a flip detector or clean-run confidence to predict future flips. It uses $U$ as an intervention subspace and gates steering with the current unedited token margin during decoding of the perturbed input (Section~\ref{sec:method}). The diagnostics therefore distinguish static prediction from decoding-time intervention; they do not by themselves establish the benefit of dynamic gating. That benefit is evaluated directly by the controlled gate ablation (Table~\ref{tab:ablation-main}). We select $(\ell,\alpha)$ using held-out validation performance rather than these proxies (Appendix~\ref{app:selection-details}).}

\FloatBarrier
\section{Additional Evaluation Results}
\label{app:additional-results}

\subsection{Recovery--preservation trade-offs}
\label{app:tradeoffs} {We plot $P=(\mathrm{C}+\mathrm{N})/2$ against flip recovery $\mathrm{F}$ for all 18 settings in Tables~\ref{tab:main-qwen} and~\ref{tab:main-gemma}. Each point represents the reported configuration, without additional tuning or interpolation. A point is Pareto-nondominated if no compared configuration has at least as high $P$ and $\mathrm{F}$ and is strictly higher in one. \method is nondominated in every setting under the reported rounded values, although other methods can remain on the frontier. Thus these comparisons show trade-offs independently of $H$, not dominance at every preservation level.}

\begin{figure}[!htbp]
\centering
\includegraphics[width=\linewidth]{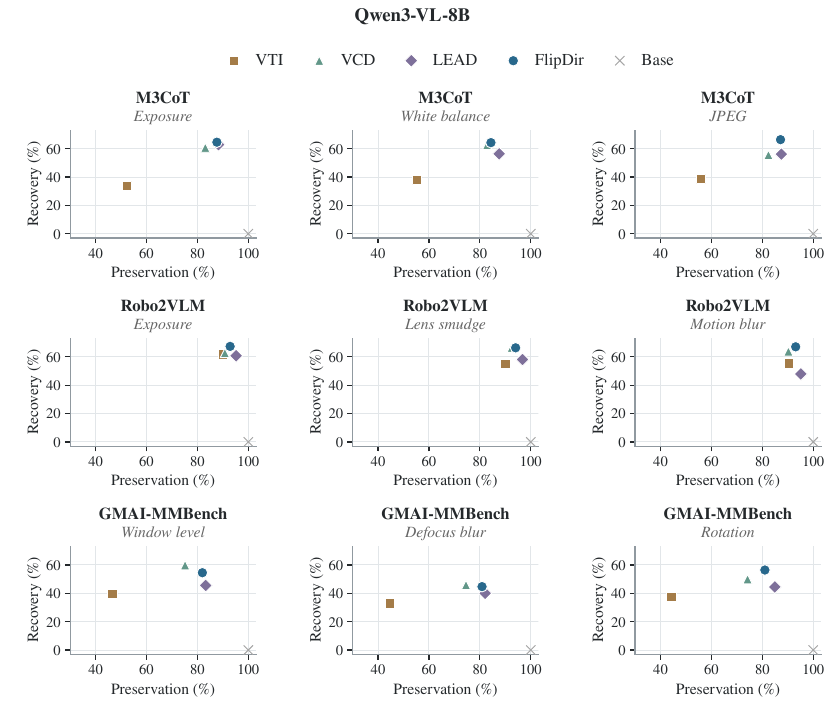}
\caption{ {\textbf{Recovery--preservation trade-offs for Qwen3-VL-8B.} All nine settings and five methods are shown on shared axes. Higher and further right indicate better recovery and preservation, respectively.}}
\label{fig:tradeoff-qwen-all}
\end{figure}

\begin{figure}[t]
\centering
\includegraphics[width=\linewidth]{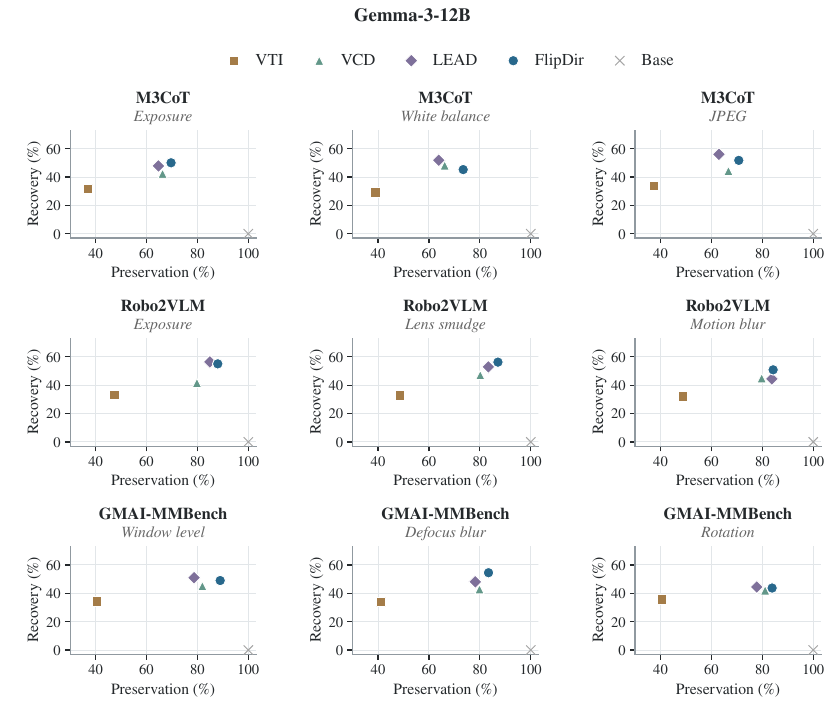}
\caption{ {\textbf{Recovery--preservation trade-offs for Gemma-3-12B.} All nine settings, with the same methods and axes as Figure~\ref{fig:tradeoff-qwen-all}. Base is at $(100,0)$ in every setting.}}
\label{fig:tradeoff-gemma-all}
\end{figure}

\FloatBarrier
\subsection{Transfer across datasets and perturbations}
\label{app:subspace-analysis}

\label{app:transfer-details}

Table~\ref{tab:transfer-full} expands Figure~\ref{fig:analysis2} with all measured source-to-target transfers, including within-domain sources. Only $U$ is transferred; the layer and strength use the target-specific configuration. Across the cross-domain combinations shown in Figure~\ref{fig:analysis2}, at least $91\%$ of target-matched $H$ is retained.

\begin{table}[!htbp]
\centering
\caption{\textbf{Transfer results.} $\hrec$ using the source subspace and the target-specific layer and strength. The last column uses a target-matched subspace.}
\label{tab:transfer-full}
\small
\begin{tabular}{ll l cc}
\toprule
Model & Target setting & Source $U$ & $\hrec$ & Target-specific \\
\midrule
Qwen3-VL-8B & Robo2VLM lens & M3CoT exposure & 75.7 & 82.5 \\
Qwen3-VL-8B & Robo2VLM lens & GMAI detector & 76.3 & 82.5 \\
Qwen3-VL-8B & M3CoT exposure & Robo2VLM lens & 71.5 & 78.3 \\
Qwen3-VL-8B & M3CoT exposure & GMAI detector & 74.2 & 78.3 \\
Qwen3-VL-8B & GMAI detector & M3CoT exposure & 65.3 & 70.7 \\
Qwen3-VL-8B & GMAI detector & Robo2VLM lens & 69.7 & 70.7 \\
Gemma-3-12B & M3CoT exposure & Robo2VLM lens & 61.5 & 61.6 \\
Gemma-3-12B & M3CoT exposure & GMAI detector & 61.4 & 61.6 \\
Gemma-3-12B & Robo2VLM lens & M3CoT exposure & 69.4 & 73.6 \\
Gemma-3-12B & Robo2VLM lens & GMAI detector & 71.6 & 73.6 \\
Gemma-3-12B & GMAI detector & M3CoT exposure & 62.3 & 64.2 \\
Gemma-3-12B & GMAI detector & Robo2VLM lens & 60.5 & 64.2 \\
\addlinespace
Gemma-3-12B & M3CoT exposure & M3CoT JPEG & 61.6 & 61.6 \\
Gemma-3-12B & M3CoT exposure & Robo2VLM exposure & 61.1 & 61.6 \\
Gemma-3-12B & Robo2VLM lens & Robo2VLM exposure & 71.1 & 73.6 \\
Gemma-3-12B & GMAI detector & GMAI window & 61.1 & 64.2 \\
Qwen3-VL-8B & GMAI detector & GMAI window & 60.9 & 70.7 \\
Gemma-3-12B & Robo2VLM lens & Robo2VLM motion & 71.6 & 73.6 \\
Gemma-3-12B & GMAI detector & Robo2VLM motion & 62.1 & 64.2 \\
Qwen3-VL-8B & Robo2VLM lens & Robo2VLM motion & 80.9 & 82.5 \\
Qwen3-VL-8B & GMAI detector & Robo2VLM motion & 70.2 & 70.7 \\
\bottomrule
\end{tabular}
\end{table}

\FloatBarrier
\section{Ablations and Sensitivity}
\label{app:ablation}

\label{app:sensitivity}

This section supplements the rank, gate, and strength analyses in Section~\ref{sec:ablations} with tabulated rank results, the full margin-gate comparison, and sensitivity to the number of calibration pairs.

\subsection{Flip-subspace rank}
\label{app:rank}
Table~\ref{tab:app-rank} sweeps $k$ on six settings, covering two non-medical perturbation families and one medical perturbation family per model, while holding all other components fixed at the selected configuration.

\begin{table}[!htbp]
\centering
\caption{\textbf{Rank sensitivity.} $\hrec$ with all other components fixed to Table~\ref{tab:app-config}. The shared rank $k=8$ is shaded; row maxima are bold.}
\label{tab:app-rank}
\small
\begin{tabular}{ll cc>{\columncolor{gray!10}}c cc}
\toprule
Model & Setting & $k = 2$ & $4$ & $8$ & $16$ & $32$ \\
\midrule
\multirow{3}{*}{Qwen3-VL-8B} & M3CoT exposure   & 71.0 & 74.3 & \textbf{78.3} & 73.2 & 72.6 \\
                             & Robo2VLM lens smudge & 82.2 & 79.6 & \textbf{82.5} & 79.1 & 77.8 \\
                             & GMAI rotation    & 65.4 & 64.7 & \textbf{70.7} & 63.7 & 69.6 \\
\midrule
\multirow{3}{*}{Gemma-3-12B} & M3CoT exposure   & 59.2 & 58.6 & \textbf{61.6} & 59.2 & 60.2 \\
                             & Robo2VLM lens smudge & \textbf{77.1} & 70.4 & 73.6 & 71.4 & 70.2 \\
                             & GMAI rotation    & 65.1 & 64.2 & 64.2 & \textbf{65.4} & 64.5 \\
\bottomrule
\end{tabular}
\end{table}

\Needspace{17\baselineskip}
\subsection{Margin-gate ablation}
\label{app:component-ablation}
\begin{figure}[H]
\centering
\includegraphics[width=\linewidth]{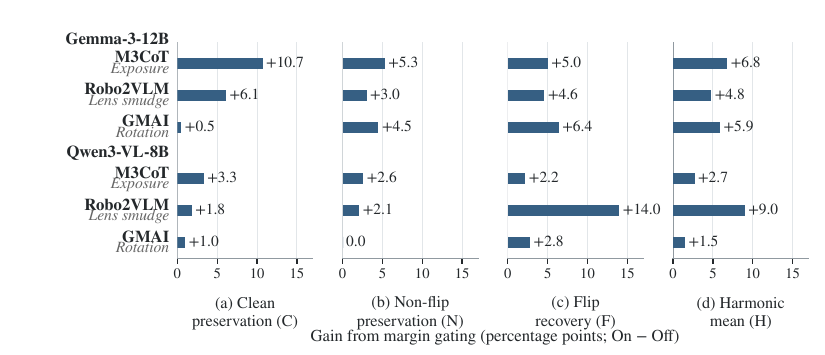}
\caption{\textbf{Performance gains from margin gating.} Changes in clean preservation (C), non-flip preservation (N), flip recovery (F), and their harmonic mean (H), measured in percentage points as Gate On minus Gate Off. All panels share the same scale; values are computed from Table~\ref{tab:ablation}. The subspace, layer, and strength are fixed within each comparison. GMAI denotes GMAI-MMBench.}
\label{fig:gate-gains}
\end{figure}

Table~\ref{tab:ablation} reports the preservation and recovery metrics underlying Table~\ref{tab:ablation-main}. Gate Off applies steering at every decoding step; the subspace, layer, and strength are held fixed within each comparison. The gate improves $H$ in all six settings. Figure~\ref{fig:gate-gains} visualizes the corresponding changes in each metric.

\begin{table}[H]
\centering
\caption{\textbf{Margin-gate ablation.} C/N/F/H (\%) with and without the gate. $\Delta H=H_{\mathrm{On}}-H_{\mathrm{Off}}$, computed from the displayed values.}
\label{tab:ablation}
\footnotesize
\setlength{\tabcolsep}{3pt}
\begin{tabular}{lrrrrrrrrr}
\toprule
& \multicolumn{4}{c}{Gate Off} & \multicolumn{4}{c}{Gate On} & \\
\cmidrule(lr){2-5}\cmidrule(lr){6-9}
Setting & C & N & F & H & C & N & F & H & $\Delta H$ \\
\midrule
\multicolumn{10}{l}{\textit{Gemma-3-12B}} \\
M3CoT exposure & 59.5 & 63.8 & 45.0 & 54.8 & 70.2 & 69.1 & 50.0 & 61.6 & +6.8 \\
Robo2VLM lens smudge & 81.7 & 83.4 & 51.6 & 68.8 & 87.8 & 86.4 & 56.2 & 73.6 & +4.8 \\
GMAI rotation & 84.0 & 78.7 & 37.3 & 58.3 & 84.5 & 83.2 & 43.7 & 64.2 & +5.9 \\
\midrule
\multicolumn{10}{l}{\textit{Qwen3-VL-8B}} \\
M3CoT exposure & 84.1 & 85.3 & 62.3 & 75.6 & 87.4 & 87.9 & 64.5 & 78.3 & +2.7 \\
Robo2VLM lens smudge & 92.3 & 91.9 & 52.3 & 73.5 & 94.1 & 94.0 & 66.3 & 82.5 & +9.0 \\
GMAI rotation & 79.2 & 81.8 & 53.6 & 69.2 & 80.2 & 81.8 & 56.4 & 70.7 & +1.5 \\
\bottomrule
\end{tabular}
\end{table}

\subsection{Calibration sample size}
\label{app:pair-count}
Table~\ref{tab:pairs} varies the number of pairs used to estimate $U$, with other components fixed. Each subset contains the first $N$ parents in sorted identifier order, with at most one flip pair per parent; the basis is estimated by uncentered SVD.

\begin{table}[H]
\centering
\caption{\textbf{Subspace-estimation sample size.} $\hrec$ with other components fixed; parent counts are shown in parentheses.}
\label{tab:pairs}
\small
\begin{tabular}{l cc c}
\toprule
Setting & $50$ pairs & subsampled & full pool \\
\midrule
Gemma M3CoT exposure   & 60.2 & 56.8 (250) & 61.6 (989) \\
Gemma Robo2VLM lens    & 67.0 & 72.2 (150) & 73.6 (319) \\
Gemma GMAI detector    & 67.0 & 63.8 (100) & 64.2 (192) \\
Qwen Robo2VLM lens     & 78.6 & 79.2 (150) & 82.5 (255) \\
\bottomrule
\end{tabular}
\end{table}

\FloatBarrier
\section{Compute, Runtime, and Gate Statistics}
\label{app:cost}

\subsection{Compute accounting}
\label{app:compute-accounting}
\paragraph{Offline direction construction.}
Direction estimation is performed once per backbone and perturbation setting, and the resulting subspace is reused across queries. Given existing clean and flip pairs, activation extraction requires two prefill-only forwards per pair; the candidate layers are read from the same forwards. The $4{,}512$ Gemma pairs and $3{,}609$ Qwen pairs in Table~\ref{tab:app-sizes} therefore require $9{,}024$ and $7{,}218$ prefills, respectively. SVD is then applied to the activation differences, and the retained basis stores $dk$ coefficients. These extraction counts exclude pair search and validation-based parameter selection. Pair-search costs are reported separately in Table~\ref{tab:construction-cost}; validation selects the steering layer and strength before deployment. Thus the method incurs an initial construction cost in exchange for a reusable intervention, without updating model weights.

\paragraph{Decoding FLOPs.}
We count a multiply-add as two FLOPs and approximate the dominant linear operations in the decoder blocks and output head. Comparisons fix the input and output lengths and exclude prefill, image encoding, and the context-length-dependent attention products. Let $F_b$ be the per-token cost of one decoder block, $F_h$ the output-head cost, and $L$ the number of blocks. Standard decoding costs $F_0=LF_b+F_h$. Under conditional execution, an open gate triggers recomputation only above the intervention layer $\ell$, including the output head. The relative cost is
\begin{equation}
\frac{F_{\mathrm{FlipDir}}}{F_0}
\approx 1+p_{\mathrm{open}}\frac{(L-\ell)F_b+F_h+4dk}{LF_b+F_h}.
\label{eq:decode-cost}
\end{equation}
The per-token cost consists of the unedited forward ($F_0$), negligible gate arithmetic, and, on open-gate steps, projection ($4dk$) plus the remaining blocks and output head ($(L-\ell)F_b+F_h$). A closed gate incurs neither projection nor recomputation. The output head is recomputed even when $\ell=L$, so intervention at the final layer does not have zero cost.

For Gemma-3-12B, $L=48$, $d=3840$, $F_b\approx4.48\times10^8$, and $F_h\approx2.01\times10^9$ FLOPs. For Qwen3-VL-8B, $L=36$, $d=4096$, $F_b\approx3.86\times10^8$, and $F_h\approx1.24\times10^9$. We combine the layers in Table~\ref{tab:app-config} with the gate-open fractions replayed on stored unedited traces in Table~\ref{tab:cost}. Estimated multipliers range from $1.068$ to $1.185$ for Gemma and $1.011$ to $1.176$ for Qwen, across seven and eight settings with available traces, respectively. These replay fractions are proxies for conditional execution, rather than measurements from a deployed conditional implementation. The unweighted mean over these 15 settings is $1.078\times$, rounded to $1.08\times$ in Table~\ref{tab:compute}.

LEAD uses one decoder forward and a dense probability-weighted embedding, $\sum_{v=1}^{V}p_t(v)e_v$, at each step. This adds approximately $2Vd$ FLOPs, giving $F_{\mathrm{LEAD}}/F_0\approx1+2Vd/F_0$: $1.086\times$ for Gemma and $1.082\times$ for Qwen. Their mean is $1.084\times$, rounded to $1.08\times$ in Table~\ref{tab:compute}. VTI adds direction updates and normalization within a single forward pass \citep{vti2024}, giving approximately $1\times$ at the scale of decoder FLOPs. VCD evaluates original- and distorted-image branches at every decoding step \citep{vcd2024}, giving approximately $2\times$; its implementation maintains separate generation states for the two branches. Under the same accounting, conditional \method uses about $41$ to $50\%$ fewer decoding FLOPs than VCD. This comparison concerns decoding arithmetic rather than end-to-end latency. The second, steered forward is executed only when the gate opens. Table~\ref{tab:compute} estimates the cost with reuse of activations below the intervention layer; a full second decoder forward instead costs approximately $1+p_{\mathrm{open}}$ times standard decoding, plus the projection. Section~\ref{app:runtime} reports its measured wall-clock runtime separately.

\Needspace{30\baselineskip}
\subsection{Gate activation statistics}
\label{app:gate-stats}
Gate open rates replay the selected thresholds $(0.3,0.7)$ with hysteresis over stored unedited traces: twelve parents per setting across shards, including clean, flip, and non-flip generations. N/A marks settings without stored traces.

\begin{table}[H]
\centering
\caption{\textbf{Runtime and gate statistics.} Time is measured in seconds per parent under the shared serving regime. Gate statistics are replayed on stored unedited traces; entries are counted per 100 tokens. N/A denotes unavailable measurements.}
\label{tab:cost}
\small
\begin{tabular}{l rr rrrr}
\toprule
& \multicolumn{2}{c}{Gate} & \multicolumn{4}{c}{Seconds per parent} \\
\cmidrule(lr){2-3}\cmidrule(lr){4-7}
Setting & Open \% & Entries & \ours & VCD & LEAD & VTI \\
\midrule
Qwen M3CoT exposure   & 15.5 &  9.1 & 254.1 & 260.5 &  92.4 & 252.2 \\
Qwen M3CoT wb         & 14.5 &  8.5 & 231.3 & 265.4 & 135.1 & 227.2 \\
Qwen M3CoT JPEG       & 12.9 &  8.0 & 232.7 & 258.7 & 134.6 & 241.4 \\
Qwen Robo2VLM exp.    & 17.3 &  9.6 & 121.8 & 109.0 &  71.9 &  44.7 \\
Qwen Robo2VLM lens    & N/A & N/A &  71.8 &  96.0 &  49.2 &  43.7 \\
Qwen Robo2VLM motion  & 18.8 & 10.2 & 112.7 & 120.4 &  70.8 &  42.1 \\
Qwen GMAI window      & 25.4 & 13.0 &  85.7 & 209.8 & 113.8 &  88.6 \\
Qwen GMAI defocus     & 24.1 & 12.2 &  74.5 & 209.7 & 114.2 &  80.5 \\
Qwen GMAI rotation    & 24.0 & 12.7 &  76.8 & 211.2 & 115.7 &  82.8 \\
Gemma M3CoT exposure  & 13.3 &  6.5 & 175.3 & 168.7 & N/A & 133.8 \\
Gemma M3CoT wb        & 11.0 &  5.9 & 171.1 & 172.5 & N/A & 121.3 \\
Gemma M3CoT JPEG      & N/A & N/A & 176.3 & 169.2 &  96.6 & 124.3 \\
Gemma Robo2VLM exp.   & 22.3 & 11.6 &  48.0 &  44.1 & N/A &  40.8 \\
Gemma Robo2VLM lens   & N/A & N/A &  36.4 &  35.5 &  17.7 &  38.3 \\
Gemma Robo2VLM motion & 24.4 & 10.8 &  22.3 &  44.0 & N/A &  53.2 \\
Gemma GMAI window     & 24.0 & 10.9 &  53.6 &  97.5 & N/A &  89.1 \\
Gemma GMAI defocus    & 21.7 & 11.8 &  49.2 &  92.6 & N/A &  74.5 \\
Gemma GMAI rotation   & 25.8 & 12.9 &  54.4 & 101.5 & N/A &  91.0 \\
\bottomrule
\end{tabular}
\end{table}

\subsection{Runtime}
\label{app:runtime}
Wall-clock time in Table~\ref{tab:cost} is averaged over the main-table runs under a shared serving regime with four to five concurrent processes per GPU. \method first computes the unedited distribution and executes the steered pass only when the gate is open. Across the six medical settings, the ratio of VCD runtime to \method runtime ranges from $1.8$ to $2.8$, with higher $H$ for \method in each setting. These measurements describe the shared serving regime; they do not isolate per-query latency or the conditional-execution FLOPs in Eq.~\ref{eq:decode-cost}. VCD adds a second image-conditioned decoding branch, whereas VTI performs direction updates within one branch. Without component-level profiling, we do not attribute their measured latency differences to prefill or implementation overhead alone.

\clearpage
\section{Qualitative Examples}
\label{app:qualitative}

\subsection{Answer-flip recovery}
\begin{figure}[H]
\centering
\begin{subfigure}[t]{0.36\linewidth}
  \includegraphics[width=\linewidth]{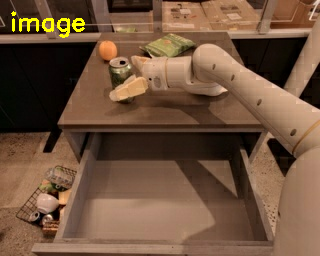}
  \caption*{\footnotesize Clean image (model answers B)}
\end{subfigure}\hfill
\begin{subfigure}[t]{0.36\linewidth}
  \includegraphics[width=\linewidth]{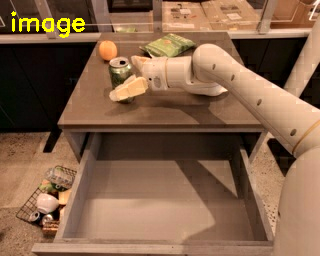}
  \caption*{\footnotesize Exposure shift ($-$EV)}
\end{subfigure}

\vspace{4pt}
{\setlength{\fboxsep}{4pt}%
\fcolorbox{black!25}{black!3}{\parbox{\dimexpr0.94\linewidth-2\fboxsep-2\fboxrule\relax}{%
\footnotesize
\textbf{[Question]} \emph{The robot is tasked to moving away with open gripper after releasing the can. After moving away with open gripper after releasing the can, what will be the robot's NEXT action phase?}\par\vspace{2pt}
\centering
(A) Moving away with open gripper after releasing the can\quad \textcolor{green!45!black}{\textbf{(B) Approaching the can with open gripper}}\\
(C) Firmly grasping the can\quad (D) Releasing the can by opening gripper\quad (E) Closing gripper to grasp the can\par}}}

\vspace{4pt}
{\setlength{\fboxsep}{4pt}\setlength{\fboxrule}{1pt}%
\begin{minipage}[t]{0.49\linewidth}
\fcolorbox{red!55!black}{red!4}{\parbox{\dimexpr\linewidth-2\fboxsep-2\fboxrule\relax}{%
\footnotesize
\textbf{Varied image, no steering} $\to$ \textcolor{red!55!black}{\textbf{(A)}}\par\vspace{2pt}
The question states that the robot has already ``Moving away with open gripper after releasing the can''. The question asks for the *next* action. Since the can has already been released, \textcolor{red!55!black}{the next logical action would be to move away with the open gripper.}\par\vspace{2pt}
FinalAnswer: (A)}}
\end{minipage}\hfill
\begin{minipage}[t]{0.49\linewidth}
\fcolorbox{green!45!black}{green!4}{\parbox{\dimexpr\linewidth-2\fboxsep-2\fboxrule\relax}{%
\footnotesize
\textbf{Varied image, \method} $\to$ \textcolor{green!45!black}{\textbf{(B)}}\par\vspace{2pt}
The question states that the robot has already ``Moving away with open gripper after releasing the can''. This means the can has been released and the robot is moving away. \textcolor{green!45!black}{The next logical action would be to approach another object or location.}\par\vspace{2pt}
Therefore, the next action phase is approaching the can with an open gripper.\par\vspace{2pt}
FinalAnswer: (B)}}
\end{minipage}}
\caption{A repaired flip on Gemma-3-12B Robo2VLM exposure shift. The unedited pass on the varied image restates the completed phase and flips to (A); with \method the reasoning advances to the cycle's next phase and the answer returns to (B), the clean answer and the ground truth. Colored spans mark where the trajectories diverge.}
\label{fig:qual-repair}
\end{figure}

Figure~\ref{fig:qual-repair} illustrates Gemma-3-12B on Robo2VLM exposure shift with $\ell=12$ and $\alpha=0.5$. 

\clearpage
\subsection{Perturbation intensity examples}
\label{app:intensity-examples}
\begin{figure}[!htbp]
\centering
\includegraphics[width=\linewidth]{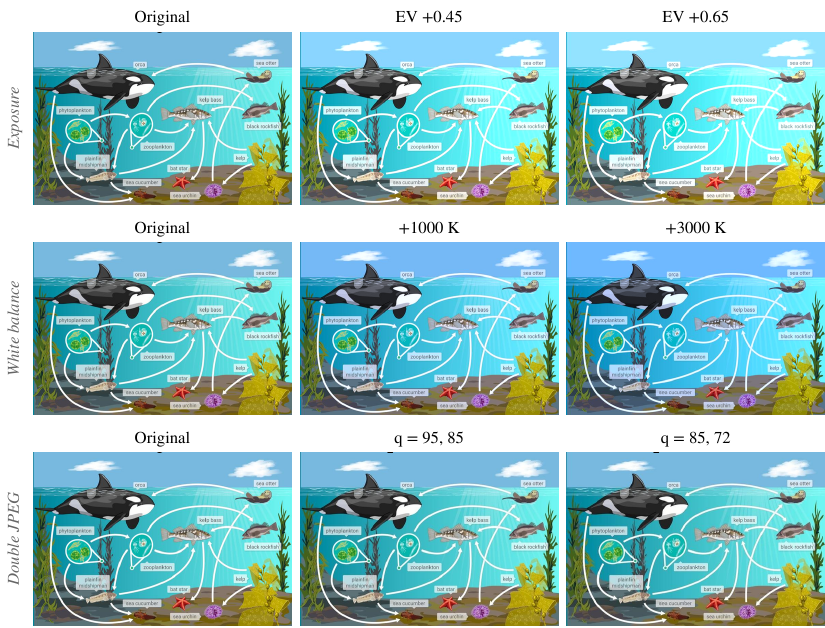}
\caption{M3CoT variation intensities on the example of Figure~\ref{fig:variations}, rendered from the fixed bands. Rows: exposure shift at the EV band endpoints ($+0.45$, $+0.65$), white-balance shift at $+1000$\,K and $+3000$\,K, and the mildest ($q = 95, 85$) and strongest ($q = 85, 72$) double-JPEG draws.}
\label{fig:app-int-m3cot}
\end{figure}

\begin{figure}[!htbp]
\centering
\begin{subfigure}[t]{0.47\linewidth}
  \includegraphics[width=\linewidth]{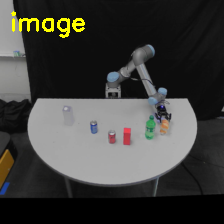}
  \caption*{\footnotesize Clean image}
\end{subfigure}\hfill
\begin{subfigure}[t]{0.47\linewidth}
  \includegraphics[width=\linewidth]{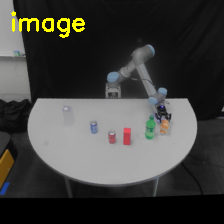}
  \caption*{\footnotesize Lens-smudge variation}
\end{subfigure}
\caption{Robo2VLM lens-smudge example from \bench. The perturbation composites wipe smears, droplets, dust, local blur, and mild haze while preserving the underlying scene.}
\label{fig:app-int-robo}
\end{figure}

\begin{figure}[!htbp]
\centering
\includegraphics[width=\linewidth]{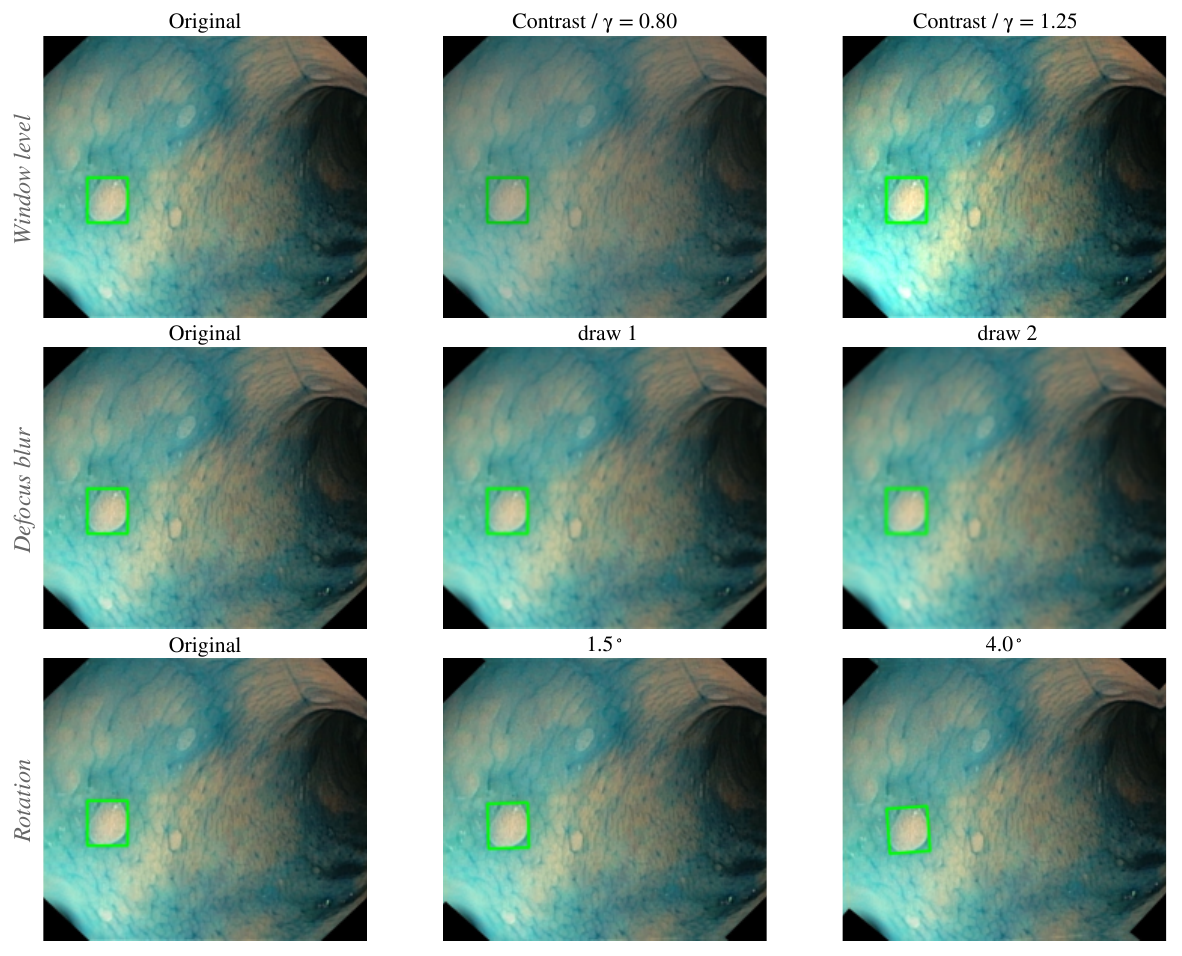}
\caption{GMAI-MMBench variation intensities on an endoscopy image. Rows: window-level shift at the joint contrast and gamma endpoints, modality-gated defocus blur (disk PSF for this lens modality), and detector rotation at $1.5^{\circ}$ and $4.0^{\circ}$.}
\label{fig:app-int-gmai}
\end{figure}

Figures~\ref{fig:app-int-m3cot}, \ref{fig:app-int-robo}, and \ref{fig:app-int-gmai} provide representative examples from the fixed perturbation bands. Additional medical-specific design details are provided in Appendix~\ref{app:medical}.

\end{document}